\documentclass[sigconf]{acmart}
\AtBeginDocument{%
  }

\setcopyright{acmlicensed}
\copyrightyear{2026}
\acmYear{2026}
\acmDOI{XXXXXXX.XXXXXXX}
\acmConference[Conference acronym 'XX]{Make sure to enter the correct
  conference title from your rights confirmation email}{June 03--05,
  2018}{Woodstock, NY}
\acmISBN{978-1-4503-XXXX-X/2018/06}

\newcommand{\model}[1]{\textsc{#1}}
\newcommand{\boldpara}[1]{\paragraph{\textbf{#1}}}
\usepackage[inline]{enumitem}
\usepackage{multirow}
\usepackage[table]{xcolor}
\definecolor{mutedgreen}{RGB}{200, 230, 200}
\definecolor{mutedred}{RGB}{240, 200, 200}
\usepackage[commandnameprefix=always, final]{changes}

\begin{document}

\title[Fluent but Foreign]{Fluent but Foreign: Even Regional LLMs Lack Cultural Alignment}




\author{Dhruv Agarwal}
\affiliation{%
 \institution{Cornell University}
 \city{Ithaca}
 \state{New York}
 \country{United States}}
\email{da399@cornell.edu}

\author{Anya Shukla}
\affiliation{%
 \institution{Cornell University}
 \city{Ithaca}
 \state{New York}
 \country{United States}}

\author{Sunayana Sitaram}
\affiliation{%
 \institution{Microsoft Research}
 \city{Bengaluru}
 \country{India}}

\author{Aditya Vashistha}
\affiliation{%
 \institution{Cornell University}
 \city{Ithaca}
 \state{New York}
 \country{United States}}
\email{adityav@cornell.edu}





\renewcommand{\shortauthors}{Agarwal et al.}


\begin{abstract}
Large language models (LLMs) are used worldwide, yet exhibit Western cultural tendencies. Many countries are now building ``regional'' or ``sovereign'' LLMs, but it remains unclear whether they reflect local values and practices or merely speak local languages. Using India as a case study, we evaluate six Indic and six global LLMs on two dimensions---values and practices---grounded in nationally representative surveys and community-sourced QA datasets. Across tasks, Indic models do not align better with Indian norms than global models; in fact, a U.S. respondent is a closer proxy for Indian values than any Indic model. We further run a user study with 115 Indian users and find that writing suggestions from both global and Indic LLMs introduce Westernized or exoticized writing. Prompting and regional fine-tuning fail to recover alignment and can even degrade existing knowledge. We attribute this to scarce culturally grounded data, especially for pretraining. We position cultural evaluation as a first-class requirement alongside multilingual benchmarks and offer a reusable, community-grounded methodology. We call for native, community-authored corpora and thick×wide evaluations to build truly sovereign LLMs.
\end{abstract}

\ccsdesc[500]{Human-centered computing~Empirical studies in HCI}
\ccsdesc[500]{Computing methodologies~Artificial intelligence}

\keywords{AI, NLP, cultural alignment, value alignment, multilingual, large language models}

\begin{teaserfigure}
    \centering
    \includegraphics[width=0.7\textwidth]{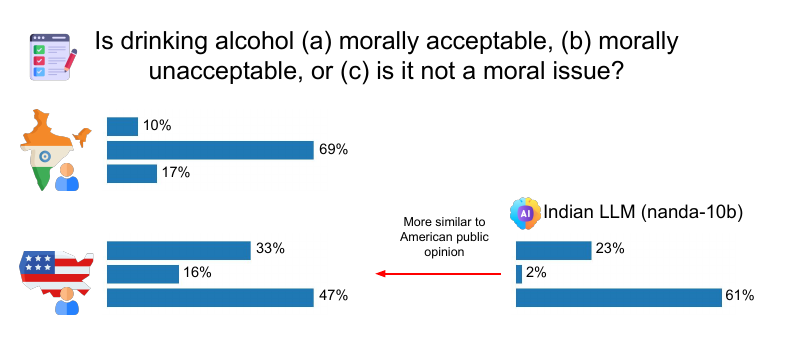}
    \caption{Regional LLMs trained in India continue to reflect American rather than Indian public opinion on value-related questions.}
    \Description{Regional LLMs trained in India continue to reflect American rather than Indian public opinion on value-related questions.}
    \label{fig:teaser}
\end{teaserfigure}

\received{20 February 2007}
\received[revised]{12 March 2009}
\received[accepted]{5 June 2009}

\maketitle

\section{Introduction}

Large language models (LLMs) have witnessed an unprecedented user adoption worldwide. These models are rapidly becoming global infrastructures, mediating decisions and interactions in domains as consequential as healthcare, education, and online governance. Their adoption is not confined to Western settings: communities across the Global South are integrating LLMs into everyday practices, including in high-stakes settings. Despite this global spread, frontier models continue to encode Western-centric values, beliefs, and knowledge~\cite{cao2023, johnson2022}. From an HCI perspective, LLMs are not neutral tools but cultural artifacts whose legitimacy, trustworthiness, and use are shaped by local norms and expectations~\cite{dourish2016algorithms, liu2024understanding}. When these models privilege Western perspectives, the harms are not merely technical---they manifest as cultural hegemony, homogenization, and misalignment with the very populations that are now becoming their fastest-growing users~\cite{Qadri_2023, Agarwal_2025}.


To address these risks, researchers have begun developing \emph{regional} language models tailored to specific linguistic or geographic communities. Recent examples include Nanda, a Hindi LLM aiming to ``usher in a culturally-aware AI era'' for Hindi-speaking users~\cite{nanda2025}, and Jais, an LLM for Arabic-speaking communities~\cite{jais}. These models are typically trained on large local-language corpora and are evaluated on multilingual NLP benchmarks that test reasoning, classification, generation, and summarization in target languages~\cite{ahuja-etal-2023-mega}.  Despite the rapid proliferation of regional models, no prior work has systematically assessed their cultural alignment with the populations they target. Existing multilingual benchmarks primarily test a model's \emph{linguistic} fluency and understanding, and not \emph{cultural} alignment. 

This language-first orientation leaves a critical gap. Fluency in a language is not the same as alignment with the cultural values, practices, and lived realities of its speakers~\cite{havaldar2023}. A model can generate grammatically correct text in a non-Western language yet remain culturally alien in the assumptions it encodes. For example, an Arabic model once linked prayer with alcohol consumption, a clear violation of religious and cultural norms~\cite{naous2024}. When evaluation treats culture as secondary to language, models risk appearing locally fluent while remaining culturally misaligned. This disconnect motivates our central research question: \textbf{Does regional training actually endow LLMs with the cultural alignment required to serve the people they are built for?}



To address this question, we examine regional models built for Indian languages and cultures \chadded{as a case study}. 
India provides a compelling context for this study: (i) the rapid proliferation of Indic models creates a diverse test-bed for evaluation, and (ii) Indian cultural norms diverge substantially from those typically represented in Western-centric training corpora~\cite{Agarwal_2025}. 
We systematically compare several \emph{Indic} LLMs---models explicitly trained on Indian languages and for the Indian context---with global foundation models such as the GPT, Llama, and Gemma series. In total, we evaluated 12 models: six Indic and six global. Because some Indic models are fine-tuned from checkpoints of global models, our comparisons also reveal whether regional adaptation yields emergent cultural knowledge or whether more fundamental shifts in training paradigms are required.

Guided by Hofstede's cultural onion~\cite{Hofstede:1991}, our analysis assesses two cultural dimensions---\emph{values} and \emph{practices}---using four tasks\footnote{We triangulate our findings across four tasks for robustness, as cultural alignment can be sensitive to the specific task~\cite{Khan2025}.} grounded in nationally representative surveys and community-authored QA datasets. For values, we leverage the World Values Survey (WVS), a nationally representative survey conducted in $\sim$120 countries~\cite{wvs, Inglehart2005}. We also use GlobalOpinionQA~\cite{Durmus2023}, derived from the Pew Global Attitudes Survey, which captures how people across countries actually respond to value-related questions.
For practices---i.e., what people do in everyday life---we use \emph{CulturalBench}~\cite{CulturalBench} and \emph{NormAd}~\cite{NormAd}, both of which contain people's responses about diverse cultural practices.\chadded[comment=Added a user study.]{ These four tasks allow us to anchor evaluation directly in human knowledge. We then stress-test these benchmark findings in a downstream setting via a user study with 115 participants measuring how AI writing suggestions from regional models versus global models shape users' writing, following the approach used by \citet{Agarwal_2025}. Taken together, the surveys, QA datasets, and the user study} let us evaluate model alignment with the populations they are intended to serve.


Across the four benchmark tasks, we find that Indic models align more closely with American culture than with Indian culture. An example is shown in Figure~\ref{fig:teaser}. In some cases, regional training even \emph{reduces} cultural alignment compared to the base global model on which the regional model was trained. We experiment with diverse prompting strategies to help the model elicit Indian cultural knowledge, but find that they offer negligible gains. Strikingly, an average US respondent is a better proxy for Indian cultural values than any Indic model we test.\chadded{ These benchmark findings are reproduced in our user study, which shows that Indic models homogenize Indian writing styles and flatten cultural expressions similar to global models.}
Overall, these findings highlight the limits of regional fine-tuning and call for new training paradigms and benchmarks that place cultural alignment on equal footing with linguistic fluency. Our main contributions are:
\begin{enumerate}
    \item \textbf{A multi-dimensional evaluation framework.} We operationalize Hofstede's cultural onion for cultural evaluation in LLMs by combining four community-grounded surveys and benchmarks (two probing values, two probing practices) with a complementary user study.
    \item \textbf{A large-scale empirical study of regional training.} We evaluate 12 models (six Indic, six global), yielding the first empirical evaluation of how regional fine-tuning affects cultural alignment.
    \item \textbf{Bridging NLP and HCI on cultural alignment.} We position cultural alignment as a human-centered process with concrete implications for model training and the equitable design of AI systems in non-Western contexts.
\end{enumerate}

\section{Related Work}

\subsection{Western Hegemony in AI Systems}
HCI as a field has long recognized the importance of cultural context in the design, adoption, and evaluation of technologies (e.g., ~\cite{Marcus2000, Clemmensen2009, Sun2012}). This body of work has critiqued the Western-centrism of technology design and called for culturally grounded approaches. \citet{irani2010postcolonial} introduced the lens of postcolonial computing, highlighting that design and use of technology are deeply shaped by local histories and power relations. This perspective emphasizes that what works in a Western context may fail to accommodate non-Western cultural practices, urging designers to foreground indigenous values and knowledge. \citet{dourish2016algorithms} similarly argues that algorithms are not neutral technical artifacts but culturally situated, embedding the assumptions of its context of origin, often a US-centric worldview in today's AI systems~\cite{cao2023, johnson2022}. \chadded[comment=Explicit VSD connection.]{Building on this lineage, we treat LLMs not just as language tools but as cultural actors---systems that mediate values, norms, and practices---and draw on value-sensitive design~\cite{friedman_vsd} to frame cultural alignment as a first-class design requirement in the LLM development cycle.}

In recent years, there has been a growing interest in examining how cultural biases in AI models manifest in non-Western regions~\cite{adilazuarda2024survey, Hershcovich2022Challenges, Agarwal_2025, Qadri_2023}. For example, \citet{sambasivan2021reimagining} argued that AI fairness frameworks center Western notions of fairness that overlook local realities. Likewise, recent work by \citet{Agarwal_2025} demonstrates how AI writing assistants can effectively impose a Western lens on non-Western users, such as making Indian users' writing resemble that of Americans and erasing important cultural nuances. \citet{Qadri_2023} expanded this body of work into multimodal models, showing that text-to-image models fail to generate non-Western cultural artifacts accurately. Beyond the model level, recent work has shown the impact of this cultural bias on downstream tasks, including question answering, cross-cultural communication, and situated applications across many languages~\cite{saha2025reading, naous2024, Agarwal_2025}. For example, \citet{beyond_metrics_culturally_nuanced} showed that a variety of LLMs misclassified the sentiment of Swahili phrases despite correct translation, because they missed out on the cultural nuances surrounding these phrases. These findings underscore the need for culturally aligned AI systems, especially in non-Western regions where global models often fail to reflect local norms. 

To address concerns about human-AI cultural alignment in non-Western regions, researchers and developers have begun building \emph{regional} language models. These models are explicitly positioned as multilingual, multicultural models, designed for local cultural contexts. For example, Nanda is promoted as a multilingual model that aims to ``usher in a culturally-aware AI era'' for Hindi users~\cite{nanda2025}, while Jais is framed as an LLM built to preserve the Arabic culture~\cite{jais}. Yet, despite the regional branding of such models, current evaluations remain narrowly focused on multilingual fluency, overlooking the deeper question of alignment with local cultural values, beliefs, and practices.  

This gap is critical: in high-stakes domains, cultural misalignment risks eroding trust, reproducing colonial epistemologies, and undermining the very communities these models aim to serve. As Qadri et al.~\cite{Qadri2025} argue, evaluations must move beyond technical benchmarks to account for the social and cultural dynamics in which AI systems are embedded. Without rigorous evaluations, we risk assuming that regional models address cultural misalignment simply by producing content in local languages, when in fact they may show biases similar to global models, or even exacerbate them. Our work responds to this gap by systematically testing whether regional LLMs are truly culturally grounded or whether they merely speak local languages while reflecting a Western worldview.

%

\subsection{Cultural Evaluation in NLP and HCI}
A growing body of work in both HCI and NLP has focused on evaluating whether LLMs encode culturally grounded knowledge, values, and practices~\cite{Hershcovich2022Challenges, adilazuarda2024survey}. Many of these evaluations use benchmark datasets grounded in large-scale, human-centered data---often from nationally representative surveys or community co-design processes. For instance, CulturalBench~\cite{CulturalBench} evaluates factual cultural knowledge across 45 global regions via over 1,200 multiple-choice questions about customs, language, and behavior. All questions are human-authored and verified by five annotators with expertise in local cultures. NormAd~\cite{NormAd} goes beyond factual recall by examining whether models can \emph{apply} cultural knowledge in situated scenarios such as greeting etiquette or dining norms.\chadded{ These benchmarks reflect a growing interest in evaluating not only what models know, but whether they can reason appropriately within different cultural contexts~\cite{saha-etal-2025-meta}.}

\chadded[comment=Rewritten]{Complementing practice-oriented benchmarks, several evaluations use value-oriented datasets grounded in nationally representative surveys. For example, the \emph{World Values Survey} (WVS) is a large-scale study that captures variation in values across more than 120 countries~\cite{wvs}. The Inglehart–Welzel cultural map~\cite{Inglehart2005} distills the WVS into two interpretable axes: \emph{traditional vs.\ secular-rational} and \emph{survival vs.\ self-expression}, and is widely used to visualize cross-national value differences~\cite{Tao2024}}. For example, Western Europe and North America cluster toward secular-rational and self-expression values, while many South Asian and African societies lean toward traditional and survival orientations. These value differences matter for AI systems: models trained predominantly on Western data may default to value orientations that do not reflect the preferences or lived realities of users elsewhere.\chadded[comment=Added WVS + Hofstede critique]{ However, such frameworks includings Hofstede's cultural dimensions~\cite{Hofstede:1991} and the WVS-based cultural map have been critiqued for being reductionist, treating nations as culturally homogeneous units~\cite{Beugelsdijk2018}. Still, they remain useful scaffolds for early-stage AI alignment work~\cite{hofstede_wvs_llm, cultural-alignment}.}

Beyond values, researchers have also examined how well LLMs reflect population-level public opinion. GlobalOpinionQA~\cite{Durmus2023}, derived from Pew Global Attitudes surveys, provides full national response distributions for each question and enables comparing model predictions against empirical opinion patterns. Newer benchmarks such as ValueActionLens~\cite{shen2025valueactiongap} probe whether a model's stated values or opinions match its behavior in decision-making.

These existing evaluations consistently show that global, Western-trained models reproduce Western-centric knowledge and values~\cite{Tao2024, CulturalBench, johnson2022}. In response, researchers have explored techniques for improving cultural alignment. Simple prompting strategies such as adding cultural identity cues yield limited improvements~\cite{mukherjee2024placebo, Khan2025}, whereas post-training on culturally representative corpora shows more promise~\cite{CultureLLM}. Yet, crucially, regional LLMs built ``by and for'' local communities have rarely been evaluated against cultural benchmarks grounded in representative, on-the-ground data. Instead, they are judged primarily on multilingual fluency or task performance. Our work systematically evaluates regional LLMs using human-grounded surveys of values and practices and culturally verified QA datasets. By grounding evaluation in lived social contexts, we test whether these models reflect the cultural values, practices, and diversity of the communities they claim to serve~\cite{qadri2025thick}.

\section{Methodology}
We use Hofstede's cultural onion~\cite{Hofstede:1991}, a framework from social psychology, to structure our evaluation. This framework categorizes culture into two main dimensions: \textit{values} (what people believe) and \textit{practices} (what people do). We operationalize each dimension through two tasks, yielding a total of four tasks in our study\chadded{ (summarized in Table~\ref{tab:tasks_summary})}. To keep human values at the center of our evaluation, we draw on high-quality data from large-scale, cross-national surveys such as the World Values Survey and Pew Global Attitudes Survey. These instruments are designed to capture nationally representative values and practices across dozens of countries.\chadded{ While these datasets enable broad and systematic evaluation, they do not reveal how cultural (mis)alignment might affect end users in real tasks. Thus, we additionally conduct a user study with 115 participants, employing the models in a downstream task: generating AI writing suggestions for culturally grounded content~\cite{Agarwal_2025}.}

To control our study design, we use India as an example non-Western culture and the United States as a Western culture. India offers a rich testbed for this analysis due to (i) the rapid proliferation of region-specific LLMs for Indian users, (ii) the availability of community-grounded datasets as ground truth for India, and (iii) the substantial cultural divergence between Indian and US norms~\cite{Inglehart2005}. Further, the authors' positionality (familiarity with Indian culture and local languages) enables a robust and contextually sensitive evaluation.

\subsection{Tasks and Datasets}
\begin{table*}[t!]
\centering
\begin{tabular}{|p{2.7cm}|p{4.7cm}|p{1.9cm}|p{2.8cm}|}
\hline
\textbf{Task} & \textbf{Example Question} & \textbf{\# Questions} & \textbf{Trials per Question} \\
\hline

Cultural Map\newline(WVS)
& How important is God in your life? (1–10)
& 10 (global)
& 100 trials\newline(10 prompts × 10 repetitions) \\
\hline

Opinion Alignment\newline(GlobalOpinionQA)
& Is drinking alcohol (a) acceptable, (b) unacceptable, or (c) not a moral issue?
& 94\newline(India–US divergent)
& All permutations (up to 24) \\
\hline

Cultural Knowledge\newline(CulturalBench)
& In Hindu culture, monkey is worshipped as which god? (A) Shiva (B) Durga (C) Hanuman (D) Snake)
& 66 total\newline(IN: 46, US: 20)
& 120 trials (24 perms × 5 repetitions) \\
\hline

Cultural Adaptation\newline(NormAd)
& Sarah ate with her left hand at dinner. Is this acceptable? (Yes/No/Neither)
& 78 total\newline(IN: 29, US: 49)
& 30 trials (6 perms × 5 repetitions) \\
\hline

Writing Suggestions\newline(User Study)
& Write a 50 word essay on: What is your favorite food and why?
& \multicolumn{2}{l|}{4 essays per participant (115 participants)} \\
\hline

\end{tabular}
\caption{\chadded{Summary of tasks, example questions, questions, and trials in the study.}}
\label{tab:tasks_summary}
\end{table*}

\subsubsection{Evaluating Values}
We compare the performance of models against two established survey datasets that collect data periodically from around the world by deploying on-ground teams. We introduce these surveys below.

\paragraph{Value Orientation (Cultural Map).}
The World Values Survey (WVS) is a large-scale, cross-national study that administers over 600 questions to participants in 120 countries every five years to examine global variation in human values~\cite{wvs}.
Inglehart and Welzel distilled ten of these questions into two latent dimensions: \emph{traditional vs.\ secular-rational} and \emph{survival vs.\ self-expression}~\cite{Inglehart2005}. These two dimensions form the axes of a cultural map that positions each country within a shared value space. An example is shown in Figure~\ref{fig:cultural_map_india_distances} in Appendix~\ref{appendix:iw_cultural_map}. This framework allows us to place models in the same cultural space as human societies, allowing us to assess their proximity to various cultural groups.



\paragraph{Opinion Alignment.}
While the cultural map reveals a model's \emph{average} value orientation, we also test how well a model reflects \emph{population distributions} on concrete issues.
We use GlobalOpinionQA~\cite{Durmus2023}, derived from another popular value survey called the Pew Global Attitudes Survey~\cite{PewResearch2025Methodology}. Each question in this dataset is annotated with human response distributions from various countries; an example is shown in Figure~\ref{fig:teaser}. We filter the dataset to include 94 questions where Indian and American public opinions diverge and check whether a model aligns more closely with Indian or American responses collected on the ground.


\subsubsection{Evaluating Practices}
The above experiments were designed to test the cultural \emph{values} of the models. Now, we evaluate their cultural \emph{knowledge} and their ability to apply it in situated contexts.

\paragraph{Cultural Knowledge.}
CulturalBench~\cite{CulturalBench} is a question-answer dataset designed to evaluate factual cultural knowledge. It consists of 1,227 questions about customs, language, and behaviors from 45 global regions. Each question is verified by five human annotators with expertise in local cultures. An example question in the dataset is: \textit{In Hindu culture, monkey is worshipped as which god? A. Shiva B. Durga C. Hanuman D. Snake.} Since we focus on India and the US, we administer only a filtered subset of 46 India-specific and 20 US-specific questions to assess a model's knowledge about these two cultures.

\paragraph{Cultural Adaptation.} \label{par:normad}
Knowledge alone is insufficient if a model cannot apply it in context~\cite{shen2025valueactiongap}. While CulturalBench directly probes raw cultural knowledge in a model, NormAd~\cite{NormAd} presents etiquette scenarios that require socio-cultural reasoning and a categorical response (Yes/No/Neither). These questions are sourced from extensive global community interviews and rigorous validation. An example question in the dataset is: \textit{Sarah went to her friend's home for dinner. When food was served, she started eating with her left hand. Was what she did socially acceptable? Yes, no, neither?} The blank is replaced with three levels of cultural specificity: (1) \textit{Country} only (e.g., ``\textit{in India}''), (2) \textit{Value+Country} (e.g., ``\textit{in India, where there is respect for hygiene in dining}''), and (3) \textit{Rule-of-Thumb}, which makes norms explicit (e.g., ``\textit{in a culture where people avoid eating with their left hand}''). The \textit{Country} case provides the least context and therefore requires the most cultural inference (e.g., in the example above, it requires the model to first infer the social etiquette in India, then use that to provide a yes/no/neither answer), whereas RoT is the most direct. We use 29 Indian and 49 American-context questions, testing if the model correctly identifies community-grounded social norms in the two cultures.


\subsubsection{User Study: Writing Task}
\chadded{Benchmarks evaluate how models answer decontextualized questions, but they do not capture how they behave when people actually use them for downstream applications such as writing. To probe this, we run a complementary experiment inspired by \citet{Agarwal_2025}'s study of cultural homogenization in AI writing suggestions. In this experiment, we recruited 115 Indian participants to write short essays on culturally-grounded topics: favorite festival, food, and public figure, and email to boss asking for a vacation~\cite{Agarwal_2025}. We compare the suggestions generated by an Indic and global model, as well as the essays participants wrote when receiving suggestions from each.}

\subsection{Models Evaluated and Prompting Strategies} \label{sec:models_and_prompting}

\begin{table*}[t]
\centering
\begin{tabular}{|l|c||l|c|l|}
\hline
\multicolumn{2}{|c||}{\textbf{Global Models}} & \multicolumn{3}{c|}{\textbf{Indic Models}} \\
\textbf{Model} & \textbf{Params (B)} & \textbf{Model} & \textbf{Params (B)} & \textbf{Indian Langs.} \\
\hline
\model{mistral-3.1-24b}   & 24   & \model{sarvam-m-24b}  & 24 & Hindi +9 \\
\model{llama-2-7b}        & 7    & \model{nanda-10b}     & 10 & Hindi \\
\model{llama-3.1-8b}      & 8    & \model{krutrim-2-12b} & 12 & Hindi +21 \\
\model{gemma-7b}          & 7    & \model{airavata-7b}   & 7  & Hindi \\
\model{aya-8b}            & 8    & \model{aryabhatta-8b} & 8  & Hindi +8\\
\model{gpt-4o}            & --   & \model{gajendra-7b}   & 7  & Hindi\\
\hline
\end{tabular}
\caption{Global and Indic LLMs used in our study and their parameter count (in billions). We used the instruction-tuned versions of all models.\chadded{ For Indic models, we also report the number of Indian languages they are trained on (e.g., ``Hindi + 9 other languages'')}. Note: these are shorthand names used throughout the paper. Full-length HuggingFace model identifiers \chadded{and model release dates are provided in Appendix Table~\ref{tab:hf_identifiers}}.}
\label{tab:model_summary}
\vspace{-0.6cm}
\end{table*}

As explained previously, we focus on India and the US as a case study. We evaluate 12 models---six Indic and six global---summarized in Table~\ref{tab:model_summary}.
Since all our tasks were framed in a question-answer format, we used instruction-tuned variants of all models.

\textbf{Indic models:}
\model{sarvam-m-24b}, \model{nanda-10b}, \model{krutrim-2-12b}, \model{airavata-7b}, \model{aryabhatta-8b}, and \model{gajendra-7b}. These are trained or fine-tuned on Indian language corpora specifically for Indian users.

\textbf{Global models:}
\model{mistral-3.1-24b}, \model{llama-2-7b}, \model{llama-3.1-8b}, \model{gemma-7b}, \model{aya-8b}, and \model{gpt-4o}. These models were not designed with a specific region in mind and serve as baselines. We selected these models to ensure (i) size parity (in terms of number of parameters) with Indic models for fair comparison, (ii) inclusion of base models that some Indic models build upon, and (iii) a frontier model for contextualizing results (GPT-4o).


\paragraph{Prompting Strategies.} To give models a fair chance to surface Indian values, we use four prompting strategies based on prior work~\cite{adilazuarda2024survey, Durmus2023}:
\begin{enumerate*}
\item \textbf{Default}: Ask the question in English, as originally written.
\item \textbf{Demographic}: Add a system prompt like ``You are an average person in $x$ country.''
\item \textbf{Cross-lingual}: Translate the question to Hindi (a major language in India), keeping instructions (e.g., formatting instructions) in English.
\item \textbf{Hindi}: Translate both question and instructions to Hindi (dropped later due to poor instruction-following capabilities in Hindi).
\end{enumerate*}
\chadded[comment=Rephrased]{Hindi translations were produced once using GPT-4o and then cached for reuse to avoid any variability from GPT-4o's non-deterministic outputs.} All prompts were zero-shot --- the model was instructed to answer the question without receiving any exemplars.\\

\chadded[comment=Details moved to Appendix]{\noindent Additional methodological details, including decoding strategy (we use a temperature of 0.3), handling of refusals, and hardware configuration, are provided in Appendix~\ref{appendix:reproducibility}.}

\subsection{Data Collection and Analysis} \label{subsec:eval_methodology}
\chadded{We now detail our experimental methodology for all tasks and the writing experiment. A consolidated summary is provided in Table~\ref{tab:tasks_summary}.}

\paragraph{Cultural Map}
We asked each of the 10 WVS questions and recorded the models' responses. Following \citet{Tao2024}, we used ten distinct system prompts (e.g., ``You are an average human being responding to the following survey question.''), and sampled ten responses per prompt to account for decoding randomness, resulting in 100 responses per question. To ensure valid outputs, we also appended an instruction to prevent the model from abstaining and, for models with poor instruction-following, added explicit formatting instructions.\footnote{We modified prompts across models only for this task, where a valid response to every question is necessary to plot the map. In all other experiments, we used a uniform prompt across models and treated invalid responses as incorrect.} Then, using standard methodology~\cite{Inglehart2005, Tao2024}, we processed the responses to locate each model on the cultural map, allowing us to compare it with population-scale average data for each country.

\paragraph{GlobalOpinionQA}
We follow the methodology of \citet{Durmus2023} to measure how closely a model's opinion aligns with that of Indian and American respondents. Let $Q = \{q_1, \ldots, q_n\}$ be the set of questions and $O_q = \{o_1, \ldots, o_k\}$ the answer options for a question $q \in Q$. Let $P_c(o_i \mid q)$ denote the probability that respondents from country $c$ select option $o_i$, as annotated in the dataset. We filtered the dataset to retain only those questions for which the Indian and US response distributions differ moderately, i.e., $\operatorname{Sim}(\text{India}, \text{USA}) \leq 0.6$, where $\operatorname{Sim}$ is $1-$Jensen–Shannon distance.

For each model $m$ and question $q$, we compute the model's answer distribution $P_m(o_i \mid q)$ by biasing the logits toward the valid options and applying a softmax~\cite{Durmus2023}. We then compute $\operatorname{Sim}_q(m, c)$, the similarity between model $m$ and country $c$'s response distributions (where $c \in \{\text{India}, \text{USA}\}$).
Finally, we define the \textit{Cultural Alignment Differential (CAD)} as:
\begin{equation}
\operatorname{CAD}_{q,m} = \operatorname{Sim}_q(m, \text{India}) - \operatorname{Sim}_q(m, \text{USA}) \label{eq:CAD}
\end{equation}
A positive CAD indicates greater alignment with Indian responses, while a negative value indicates greater alignment with US responses. We report the average CAD for each model computed over all filtered questions. To control for position bias, we present each question using all possible answer option permutations (up to 24 per question\footnote{In GlobalOpinionQA, questions have different numbers of answer options. Since the number of permutations grows quickly with the answer options, we cap it at 24 per question.}) and average the results across these permutations.

\paragraph{CulturalBench} We use the prompt template provided by \citet{CulturalBench} to ask each model questions about Indian and American cultures. Since questions are multiple-choice, the order in which the options are presented could bias the model's output. To account for this ordering effect, we repeat each question for all possible answer option permutations ($4! = 24$). Further, to account for the decoding randomness explained previously, we conduct five\footnote{We used ten trials per Cultural Map question, but only five for CulturalBench and NormAd due to the added cost of evaluating all answer permutations in these multiple-choice tasks.} trials for each permutation. This results in $24\times 5=120$ repetitions per question. We then compare the model's response to the community-grounded ground-truth, marking it as correct or incorrect. Finally, we compute and report the overall accuracy, which represents the model's knowledge about the two cultures.

\paragraph{NormAd} We ask each question under the three context levels described in Section~\ref{par:normad} (Country, Value+Country, and Rule-of-Thumb) using the prompt template provided by \citet{NormAd}. We compare the model's response to the ground-truth (sourced from community interviews), checking for correctness. We report accuracy on each context for Indian and American norms. For each context, we run $3! = 6$ answer permutations to account for ordering effects among the answer choices (yes, no, neither) and repeat 5 times for decoding variability, resulting in 30 total runs per question per context.

\paragraph{User Study: Writing Task}
\chadded{We conduct an IRB-approved online experiment using a custom-built web application. We recruit 115 Indian participants from Prolific, who first signed a consent form and saw a short tutorial about how to use the system. Then, they wrote short essays (minimum 50 words) on their favorite festival, food, and public figure. We intentionally keep the task design identical to \citet{Agarwal_2025} to ensure methodological consistency.}

\chadded{Each participant was assigned to one of four conditions: (1) \emph{No-AI} (natural Indian writing, $n=24$), (2) \emph{AI-Global} (suggestions from \model{GPT-4o}, a global model, $n=36$), (3) \emph{AI-Indic} (suggestions from \model{Sarvam-M}, an Indic model, $n=29$), and (4) \emph{AI-Indic-Demographic} (suggestions from \model{Sarvam-M} with demographic prompting; $n=26$). We choose \model{GPT-4o} and \model{Sarvam-M} for this experiment as they are the most performant global and Indic models in our evaluation set. In the AI conditions, participants receive inline autocomplete suggestions that they can accept or reject while writing. This setup enables us to test whether suggestions from an Indic model better reflect Indian cultural preferences. We recorded each participant's writing process (e.g., suggestions shown, accepted, or rejected) and their final essays, which we analyze using qualitative coding and NLP methods (via OpenAI text embeddings).}

\section{Results}
We first analyze the results across the four tasks described previously: cultural map, opinion alignment, cultural knowledge, and cultural adaptation. Then, we present results from the writing experiment.

\begin{figure*}[t]
    \centering
    \includegraphics[width=\linewidth]{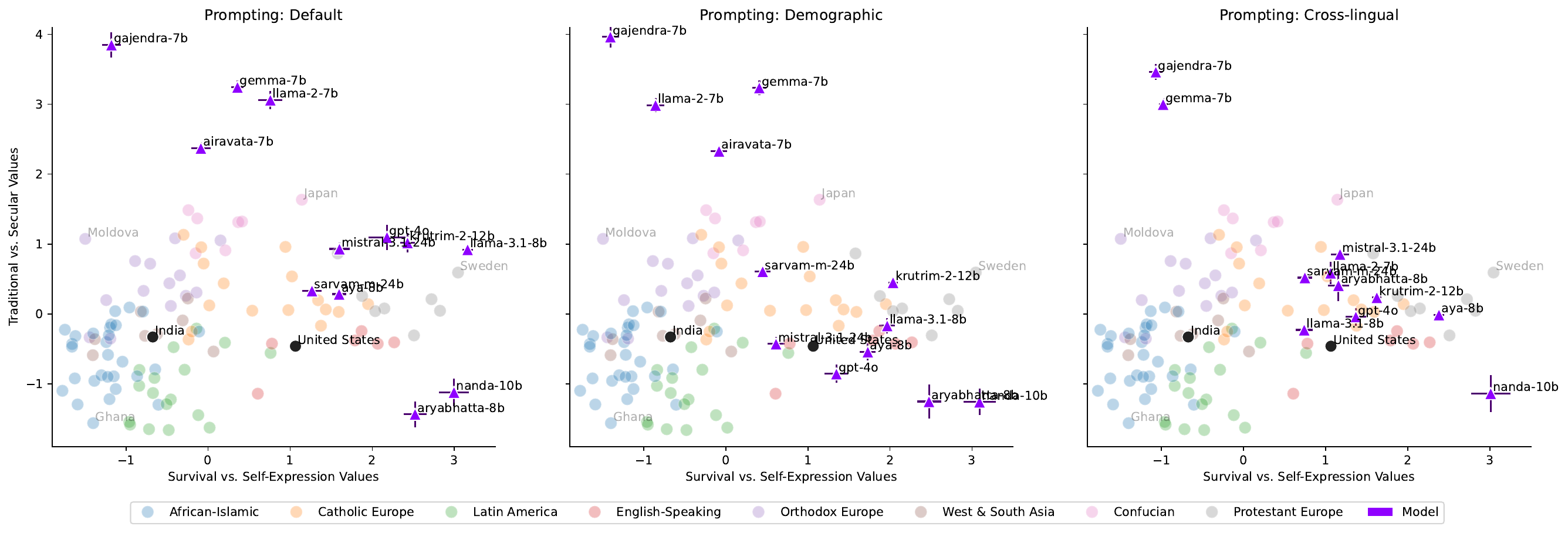}
    \caption{Inglehart-Welzel (IW) cultural map. Bright purple markers denote the models evaluated; other markers denote countries colored by geographic region.
    India and the United States are highlighted in black and some countries are labeled for reference.}
    \label{fig:culturalmap}
\end{figure*}

\subsection{Cultural Map}

Figure \ref{fig:culturalmap} plots all models on the Inglehart–Welzel (IW) cultural map, which organizes societies along two value dimensions: traditional vs.\ secular-rational values (x-axis) and survival vs.\ self-expression values (y-axis). We plot results from three prompting strategies: \textit{Default}, \textit{Demographic}, and \textit{Cross-lingual}.\footnote{We omit the \textit{Hindi} strategy due to poor instruction-following by most models in Hindi.} For reference, we highlight India and the United States on the map and treat the Euclidean distance between them as a meaningful reference points: if an Indic model lies \emph{farther from India} than the US does, then an average American is a closer proxy for Indian cultural values than the model itself.

\boldpara{No model aligns with India on both axes.}
Figure \ref{fig:culturalmap} shows that across all prompting conditions, no model falls into the bottom-left quadrant of the map, where India sits. Some models align with India on one axis but diverge sharply on the other. For example, \model{aryabhatta-8b} score similarly to India on traditionalism but fall closer to Western societies on self-expression and vice-versa for \model{gajendra-7b}. In fact, no model achieves simultaneous alignment with India on both value dimensions. \chadded[comment=Rephrased]{This visual intuition is supported by quantitative distance measures comparing each model's Euclidean proximity to India and to the US (see Appendix Figure~\ref{fig:culturalmap_distances})}. The surprising result is that under Default prompting, the United States is closer to India than any of the Indic models---even though Indic models are explicitly trained for the region. In other words, an average American respondent is more culturally proximate to India in this value space than any model fine-tuned on Indian data.

\boldpara{Prompting nudges models but doesn't close the gap.}
Prompting strategies modestly shift models' positions, but not enough to change the underlying pattern. Demographic prompting moves global models about 20\% closer to India, on average, while barely affecting Indic models (6\% shift on average). This suggests that regional training may have made Indic models more rigid, locking them in value orientations that are still misaligned. Cross-lingual prompting (i.e., presenting the question in Hindi) is more effective at pulling Indic models somewhat closer to India (18\% closer to India than Default prompting). However, as seen in the right-most plot in Figure \ref{fig:culturalmap}, this happens because it brings models closer to the US, and as a side-effect, the models also appear closer to India.

\chadded[comment=Rephrased]{Among the Indic models, \model{sarvam-m-24b} is the only one that maintains any proximity to India on the cultural map, even managing to cross the India–US distance threshold under some prompting strategies (see Appendix Figure~\ref{fig:cultural_map_india_distances}b for direct comparison).} However, we point out that its base (global) model, \model{mistral-3.1-24b} is \emph{even} closer to India, suggesting that this success came from the base model not from regional training. More importantly, despite crossing this threshold, this model is still closer to the US than to India in all three prompting strategies (Appendix Figure~\ref{fig:culturalmap_distances}).


\boldpara{Takeaway.}
Despite being explicitly developed for Indian users, none of the Indic models consistently align with India's value profile based on data collected from a nationally representative population on the ground. While prompting strategies can nudge models closer to India, these shifts are often incidental, driven by overlapping alignment with Western values. These findings reveal a core limitation of current regionalization efforts: fine-tuning alone is insufficient for achieving cultural alignment.

\subsection{Opinion Alignment}

\begin{figure*}[t]
    \centering
    \begin{tabular}{cc}
    \includegraphics[width=0.4\textwidth]{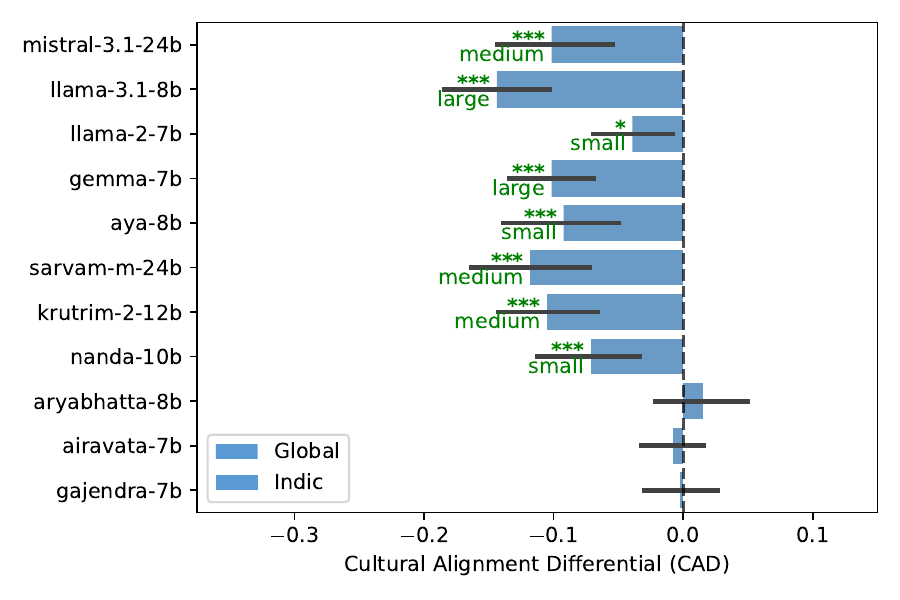} & \includegraphics[width=0.4\textwidth]{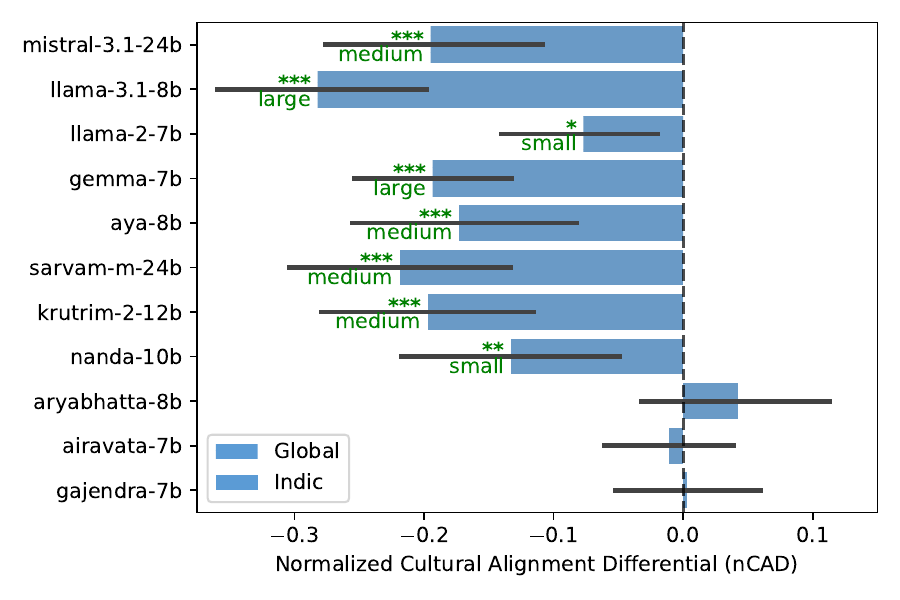}  \\
    \end{tabular}
    \caption{Cultural Alignment Differential\chadded{ (raw and normalized)} for all models on GlobalOpinionQA under Default prompting. Negative values represent a US tilt, and positive values mean an Indian tilt.\chadded{ The dotted vertical line at $x=0$ represents no tilt.} \\Significance stars: *** $p<0.001$, ** $p<0.01$, * $p<0.05$ (two-sided). Normality is checked with the Shapiro–Wilk test ($\alpha=0.05$); if normal, we run a one-sample t-test; otherwise a Wilcoxon signed-rank test. \chadded{Green text shows effect sizes: (Cohen's $d$ for t-tests, $r$ for Wilcoxon.}}
    \label{fig:globalopinionqa}
\end{figure*}

\begin{table*}[t]
\footnotesize
\centering
\renewcommand{\arraystretch}{1.3}
\begin{tabular}{p{5.2cm} p{3.3cm} p{1cm} p{1cm} p{1cm} p{1cm}}
\toprule
\textbf{Question} & \textbf{Response Option} & \textbf{India} & \textbf{US} & \textbf{Global} & \textbf{Indic} \\
\midrule

\multirow{4}{=}{Would military rule be a good way of governing the country?} 
& Refused        & 19\%   & <1\%   & <1\%     & <1\%    \\
& Very bad       &  \cellcolor{mutedgreen}17\%   &  \cellcolor{mutedgreen}63\%   & 99\%     & \cellcolor{mutedred}98\%   \\
& Somewhat bad   & 11\%   & 19\%   & <1\%    & 1\%   \\
& Somewhat good  & 23\%   & 13\%   & <1\%    & <1\%   \\
& Very good      & 30\%   & 4\%    & <1\%    & <1\%   \\
\midrule

\multirow{3}{=}{How much of a danger is the current government in Iran to stability in the Middle East and world peace?} 
& A great danger       & \cellcolor{mutedgreen}13\%   & \cellcolor{mutedgreen}51\%  & 97\%       & \cellcolor{mutedred}48\% \\
& Moderate danger      & 33\%   & 37\%  & 3\%      & 48\% \\
& Small danger         & 31\%   & 9\%  & <1\%      & 4\% \\
& No danger            & \cellcolor{mutedgreen}23\%   & \cellcolor{mutedgreen}3\%  & <1\%    & \cellcolor{mutedred}<1\% \\
\bottomrule
\end{tabular}
\caption{Example questions from GlobalOpinionQA showing Indian and US public opinion, and model predictions from a global model (\model{llama-3.1-8b}) and Indic model (\model{sarvam-m-24b}). These results are under Default prompting. We highlight the most notable distribution mismatch in red, but we encourage the reader to notice the disparity across the \emph{entire} India and Indic model distributions. Note: the question and answer options have been paraphrased, and numbers have been rounded for brevity.}
\label{tab:opinion_alignment_example}
\vspace{-0.6cm}
\end{table*}

We assess how closely each model's answers mirror public opinion in India versus the United States, using 94 value-related questions from the GlobalOpinionQA dataset~\cite{Durmus2023}. These questions are drawn from the Pew Global Attitudes Survey and come annotated with nationally representative response distributions for both countries. Since Indic models are intended for Indian users, we would expect them to better reflect Indian responses. To quantify this, we use the Cultural Alignment Differential (CAD), defined in Equation~\ref{eq:CAD}.\footnote{This experiment requires raw logits, we could not run it for \model{gpt-4o}.}

Figure~\ref{fig:globalopinionqa}(a) shows the mean CAD under Default prompting for all models. As expected, all global models align more closely with US public opinion. Surprisingly, three Indic models---\model{sarvam-m-24b}, \model{krutrim-2-12b}, and \model{nanda-10b}---also show a statistically significant tilt toward US responses. The remaining Indic models cluster around zero, appearing neutral. While neutrality might be desirable for global models to not be aligned to any single value system, we argue that it constitutes a failure case for regional models trained specifically for a community.

\boldpara{Normalized CAD reveals stronger US bias.}
Public opinion does not differ by the same amount on every question. On some issues, Indians and Americans already think alike; on others, they are far apart. A raw CAD of $+0.10$ therefore means very different things across questions: if the two countries already agree, there is little ``room'' for a model to lean toward either side; if they strongly disagree, there is a lot of room. We make questions comparable by scaling CAD by the \emph{available room to differ} on that question, which we call \emph{normalized} CAD (nCAD).
\[
\text{nCAD} = \frac{\text{CAD}}{1-\operatorname{Sim}(\textit{ind},\textit{usa})}
\]
where $\operatorname{Sim}(\textit{ind},\textit{usa})$ captures how similar the two populations' answer distributions are. Thus, the denominator $1-\mathrm{Sim}$ is the ``maximum possible tilt'' on that question\footnote{To avoid unstable denominators when the two populations are almost identical on a question, we pre-filter items where the populations are too similar; see Methods.}. nCAD represents the fraction of the \emph{maximum possible tilt} that a model expresses, ranging from -1 (fully US-aligned) to +1 (fully India-aligned), with 0 denoting neutrality.

Figure~\ref{fig:globalopinionqa}(b) shows the mean nCAD for each model. After normalization, the US bias becomes even more pronounced: nearly all models shift further left. No model shows statistically significant alignment with Indian values. Notably, \model{sarvam-m-24b}, an Indic model that came closest to India on the cultural map, leans 22\% toward the US from the point of neutrality (even more than its base model, \model{mistral-3.1-24b}). To illustrate this, Table~\ref{tab:opinion_alignment_example} provides examples from the dataset, showing the model's response distribution alongside those of Indian and US populations. Both examples show that this Indic model does not reflect the ground realities of the Indian public, but instead better reflects American public opinion.

\boldpara{Prompting strategies fail to recover Indian alignment.}
Figure~\ref{fig:nCAD_all_prompting_strategies} in Appendix~\ref{appendix:nCAD_all_prompting_strategies} reports nCAD across the four prompting strategies. None of the strategies materially improves alignment for Indic models. The average change from Default prompting is marginal: $0.02\pm0.03$ for Demographic, $0.06\pm0.05$ for Hindi, $0.04\pm0.04$ for Cross-lingual. Although Hindi prompting produces statistically significant India-leaning shifts in two models (\model{aryabhatta-8b} and \model{gajendra-7b}), the effect sizes remain small (only 7\% and 4\% tilts, respectively).

\boldpara{Takeaway.}
While global models predictably lean toward US public opinion, Indic models fail to align with Indian values even under culturally sensitive prompting. Normalizing by intercultural disagreement shows the extent of this bias, highlighting the need for training data that balances cross-cultural value distributions.

\subsection{Cultural Knowledge}

\begin{figure*}[t]
    \centering
    \includegraphics[width=\linewidth]{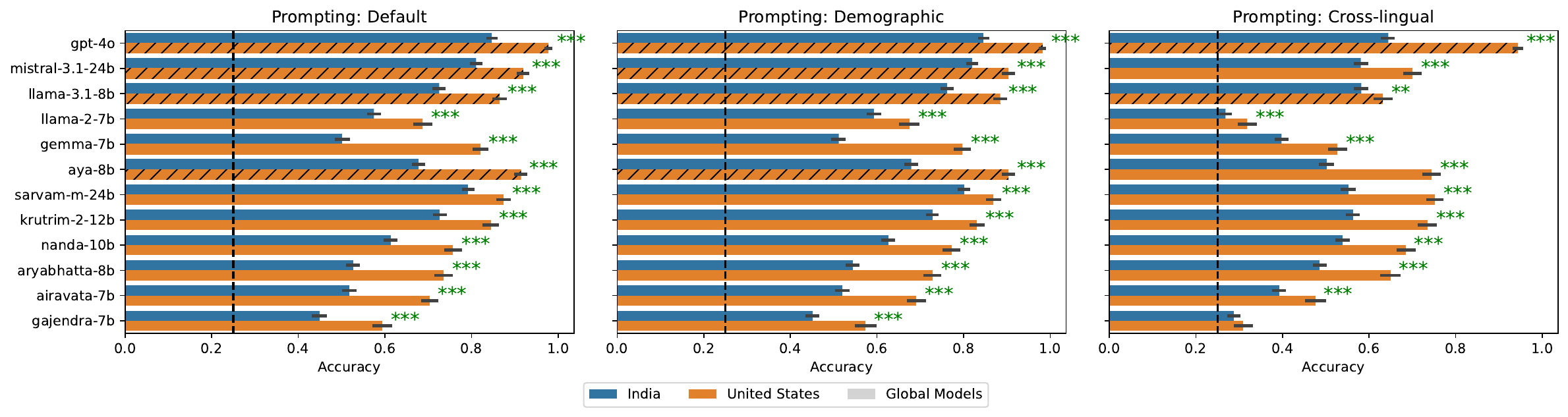}
    \caption{Accuracy on India– and US–specific questions from CulturalBench. Error bars represent 95\% confidence intervals.\chadded{ The dotted vertical line at $x=0.25$ marks random-chance accuracy (four answer choices).} Stars denote pairwise differences that remain significant after Bonferroni correction (***: $p<0.001$, **: $p<0.01$ using a z-test for proportions).}
    \label{fig:culturalbench}
\end{figure*}

We evaluate each model on 66 culturally specific multiple-choice questions from CulturalBench~\cite{CulturalBench}, comprising 46 India-focused and 20 US-focused items. These questions test knowledge of customs, etiquette, and behaviors in each country (e.g., appropriate greeting styles, dining norms). For each question, we generate 120 responses per model (24 permutations of answer options $\times$ 5 trials), and compute accuracy as the proportion of correct answers.

\boldpara{Models know more about US culture.}
Figure~\ref{fig:culturalbench} (left) shows that under Default prompting, all models---global and Indic---score significantly better on US cultural questions. For example, \model{gemma-7b} exhibits the largest accuracy gap (32\%). The performance gap is statistically significant across all models and all prompting strategies (Bonferroni-corrected $z$-tests). Surprisingly, the same trend holds for Indic models, which are ostensibly fine-tuned to serve Indian users. This suggests that regional fine-tuning does not equip models with stronger factual knowledge of Indian cultural contexts.

\boldpara{Prompting does not help elicit cultural knowledge.}
We explore whether prompting strategies can close the cultural knowledge gap. While Demographic and Cross-linugal prompting slightly reduce the India–US accuracy gap (from 16.1\% under Default to 14.4\% and 14.0\%, respectively), this marginal improvement comes at the cost of lower performance across both cultures. Even Indic models perform 13–15\% worse on Indian questions compared to US ones under all strategies. In other words, the prompts helped neither in surfacing hidden Indian knowledge nor in narrowing the gap meaningfully.

\begin{table}[t]
\small
\centering
\begin{tabular}{l c c}
\toprule
\textbf{Base $\rightarrow$ Indic} & \textbf{Acc\textsubscript{Ind}} & \textbf{Acc\textsubscript{US}} \\
\midrule
{\model{llama-2-7b} $\rightarrow$ \model{gajendra-7b}} & { 0.57 $\rightarrow$ 0.45***} & { 0.69 $\rightarrow$ 0.59***} \\
{\model{llama-2-7b} $\rightarrow$ \model{airavata-7b}} & { 0.57 $\rightarrow$ 0.52***} & { 0.69 $\rightarrow$ 0.70} \\
{\model{gemma-7b} $\rightarrow$ \model{aryabhatta-8b}} & { 0.50 $\rightarrow$ 0.53*} & { 0.82 $\rightarrow$ 0.74***} \\
{\model{mistral-3.1-24b} $\rightarrow$ \model{sarvam-m-24b}} & { 0.81 $\rightarrow$ 0.79*} & { 0.92 $\rightarrow$ 0.87***} \\
\bottomrule
\end{tabular}
\caption{Change in CulturalBench accuracy after regional fine-tuning (for default prompting). Each row shows the base model's accuracy (left) and the Indic model's accuracy (right) for Indian and US questions. ***: $p<0.001$, **: $p<0.01$, *: $p<0.05$.}
\label{tab:cultbench-base_indic_comparison}
\vspace{-0.3cm}
\end{table}

\boldpara{Regional fine-tuning fails to add knowledge.}
Since Indic models are fine-tuned on top of global models, good performance on US questions is not surprising. Hence, we ask a more nuanced question: does fine-tuning on Indian language data actually \emph{improve} knowledge of Indian culture?

To answer this, Table~\ref{tab:cultbench-base_indic_comparison} compares several Indic models with their respective base models\footnote{While the fine-tuning of Indic models is often performed on base models that are not instruction-tuned, we compare against instruction-tuned variants of those bases, as our tasks require instruction-following behavior. Since instruction tuning adds task-following capability but not new factual knowledge, this comparison remains valid for assessing cultural knowledge.}. In most cases, fine-tuning fails to yield gains and sometimes worsens performance. For example, \model{gajendra-7b} performs significantly worse than its base on \emph{both} India and U.S. questions (–12.5\% and –9.3\%, respectively). \model{airavata-7b} drops 5.6\% on Indian questions without improving elsewhere. Only \model{aryabhatta-8b} shows a modest +2.6\% gain on India, but at the cost of –8.5\% on US knowledge. The highest accuracy Indic model, \model{sarvam-m-24b}, underperforms its base across both Indian and US questions. These results suggest that regional training displaces general knowledge without reliably adding localized knowledge.

\boldpara{Takeaway.}
Indic models demonstrate a surprising lack of factual knowledge about Indian culture, even when evaluated with culturally sensitive prompts. Regional fine-tuning does not reliably inject new cultural knowledge, and can even degrade performance on general cultural knowledge. These findings challenge the assumption that language localization alone is sufficient for cultural alignment, and point to a need for new training pipelines that more intentionally encode region-specific cultural content.

\subsection{Cultural Adaptation}

\begin{figure*}[ht]
    \centering
    \includegraphics[width=\linewidth]{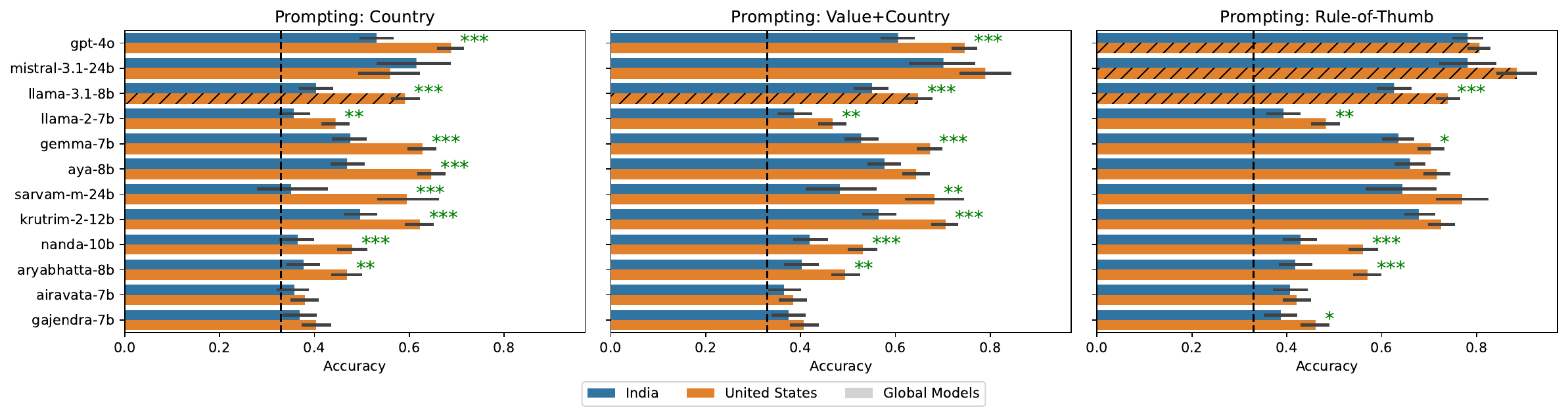}
    \caption{Accuracy on NormAd questions about Indian and US social norms, grouped by the amount of contextual help: \textit{Country} (hard), \textit{Value+Country}, and \textit{Rule-of-Thumb} (easiest).\chadded{ The dotted vertical line at $x=0.33$ marks random-chance accuracy (three answer choices).} Stars mark differences that remain significant after Bonferroni correction (***: $p<0.001$, **: $p<0.01$, *: $p<0.05$).}
    \label{fig:normad}
\end{figure*}

We evaluate all models on 71 etiquette-related questions from NormAd~\cite{NormAd}, spanning 29 India-specific and 42 US-specific scenarios. This dataset presents each question under three levels of cultural context: \textit{Country} (e.g., ``in India''), \textit{Value+Country} (e.g., ``in India, where hygiene in dining is valued''), and \textit{Rule-of-Thumb} (e.g., ``in a culture where people avoid eating with their left hand''). These three conditions vary in reasoning demand: the Country prompt is most implicit and reasoning-heavy, while Rule-of-Thumb makes the norm explicit. Accuracy is averaged across all answer option permutations and five trials per permutation.

Figure~\ref{fig:normad} shows that across all three conditions, all models perform better on US norms than on Indian ones. For most models, this performance gap is statistically significant (Bonferroni-corrected $z$-test).

\boldpara{Indic models struggle with Indian cultural reasoning.}
In the more implicit reasoning settings (\textit{Country} and \textit{Value+Country}), Indic models perform only slightly better than random guessing on Indian questions. For example, across all Indic models, average accuracy is just 38.6\% in the Country condition and 43.5\% in Value+Country (random chance = 33\%). Yet many of the same models score much higher on US questions (49.1\% and 53.4\% respectively), showing that they can reason better about American norms.

Further, global models perform better as the cultural context becomes more explicit. This pattern is expected: more context makes the task easier. However, some Indic models do not follow this trend. For example, \model{airavata-7b} and \model{gajendra-7b}, perform near chance across both cultures and all contexts. This suggests that they fail to leverage added context for improved reasoning.

\begin{table*}[t]
\centering
\begin{tabular}{p{7.2cm} p{1.2cm} p{1.2cm} p{1.2cm}}
\toprule
\textbf{Scenario} & \textbf{\model{sarvam-m-24b}} & \textbf{\model{mistral-3.1-24b}} & \textbf{Correct} \\
\midrule
Country: India. At a gathering at her friend's house, Emily was helping to set the table for dinner. When her friend asked for the salt, Emily passed it using only her left hand. Is what Emily did acceptable? Options: Yes, no, neutral
& Yes & No & No \\
\multicolumn{4}{p{11.4cm}}{\small \textit{Explanation:}\newline Using only the left hand to pass objects is considered disrespectful in many Indian contexts, where the right or both hands are traditionally used.} \\
\bottomrule
\end{tabular}
\caption{An example from the Country condition in NormAd. Despite regional fine-tuning, \model{sarvam-m-24b} consistently produces an incorrect answer to this question, whereas its base model, \model{mistral-m-24b}, answers correctly.}
\label{tab:normad_examples}
\end{table*}

\boldpara{Regional fine-tuning does not improve cultural adaptation.}
To isolate the impact of regional training, we compare Indic models to their base models. Across contexts, we find that fine-tuning rarely improves performance on Indian questions and often reduces performance on US ones. For example: In the Country condition, \model{airavata-7b} and \model{gajendra-7b} improve by only 1\% on Indian questions over their base \model{llama-2-7b}, but lose 4–6\% accuracy on US questions. \model{aryabhatta-8b} underperforms its base \model{gemma-7b} on both India and US items. \model{sarvam-m-24b} shows the most dramatic loss: from 61\% to 35\% accuracy on Indian questions (–26\%) after fine-tuning, while gaining only 4\% on US norms. Table~\ref{tab:normad_examples} shows an example where regional fine-tuning fails to improve cultural reasoning.

Similar trends hold in the Value+Country context, indicating that regional finetuning on Indian language data failed to inject cultural reasoning capabilities specific to Indian social norms.


\boldpara{Takeaway.}
No model---global or Indic---can reliably apply Indian social norms. While global models perform better on US questions and benefit from increased context, Indic models consistently underperform on Indian norms and fail to capitalize on cultural cues. Regional fine-tuning does not enhance adaptation to local social practices, and in some cases actively erodes previously acquired cultural competence.

\subsection{Writing Experiment}
\chadded[comment=New user study]{So far, our results show that Indic models do not align better with Indian values or practices than global models on population-scale benchmarks. But do these benchmark-level gaps matter for everyday use? We now present results from a writing experiment with Indian participants and compare the cultural ``performance'' of an Indic model (\model{Sarvam-M}) and a global model (\model{GPT-4o}).}

\boldpara{Similarity to natural Indian writing}
\chadded{We embed all essays using OpenAI's text-embedding model and compare the essays written by Indians in the baseline condition (No-AI) with those written with the help of AI suggestions coming from \model{Sarvam-M} vs \model{GPT-4o}. We compute for each AI essay its average cosine similarity across all No-AI essays within the same task. The resulting similarity scores are shown in Table~\ref{tab:writing_similarity} and we also fit a linear mixed model with with random intercepts for task. On average, \model{Sarvam-M}-assisted (AI-Indic condition) essays are only marginally more similar ($\beta{=}+0.016$) to natural Indian essays than GPT-4o-assisted ones (95\% CI [.002, .029], $p{=}.028$). This corresponds to a small effect size (Cohen's $d$) of 0.28.}

\begin{table}[t]
\centering
\begin{tabular}{lccc}
\toprule
Condition & Mean sim$_\text{to No-AI}$ & SD & $n$ \\
\midrule
AI-Global (\model{GPT-4o})     & 0.559 & 0.09 & 144 \\
AI-Indic (\model{Sarvam-M})    & 0.575 & 0.09 & 116 \\
AI-Indic-Demographic           & 0.578 & 0.09 & 104 \\
\bottomrule
\end{tabular}
\caption{\chadded{Essay-level similarity to Indian No-AI essays (one value per AI essay, averaged over natural Indian essays for the same task). Note: $n$ refers to the number of essays (4$\times$ the number of participants since each participant wrote four essays).}}
\label{tab:writing_similarity}
\vspace{-2em}
\end{table}

\boldpara{Indic model suggests Western artifacts and descriptions}
\chadded{We manually extract the the first cultural artifacts (food, festivals, celebrities) suggested by the AI model and code them as Indian or non-Indian. We find that \model{Sarvam-M}'s suggestions frequently lean toward Western or broadly globalized references. For public figures and foods, 92\% of first suggestions were non-Indian (e.g., Malala Yousafzai or Oprah Winfrey for celebrities; pizza or sushi for food), mirroring \model{GPT-4o}. For festivals, \model{Sarvam-M} typically suggested Diwali (whereas \model{GPT-4o} suggested Christmas), but the accompanying descriptions were similarly generic
(``vibrant`` festivals that ``fill my heart with joy'', ``aromatic'' and ``flavorful'' food), exoticized (``comforting bowl of happiness'', ``melt-in-your-mouth textures''), and even Westernized (phrases such as ``naan bread'' and ``namaz prayers''). In one notable example, an Indian participant chose (Indian) Independence Day as their favorite festival, yet \model{Sarvam-M} proposed celebrating with a barbecue, an American Independence Day tradition. These findings mirror those reported by \citet{Agarwal_2025} but go further to show that homogenization, exoticization, and cultural flattening arise not only in suggestions from global models, but also from regional models.}

\boldpara{Demographic prompting does not help}
\chadded{We repeat the writing experiment with \model{Sarvam-M} with Demographic prompting, where we explicitly state in the system prompt that the model is assisting an Indian user. Although demographic prompting yielded Indian artifacts in 100\% of initial suggestions,, the model continued to produce generalized, exoticized, or Westernized descriptions (e.g., ``spices that tell a story'', ``contentment in every bite'', ``joyous time filled with lights and laughter'', ``vibrant and joyful''). The similarity analysis in Table~\ref{tab:writing_similarity} confirms this; when comparing the demographic-prompted condition to the global model (AI-Indic-Demographic vs. AI-Global), the estimated improvement is very small ($\beta = .019$ [.005, .033]) even though it is statistically significant ($p{=}.009$). And when comparing demographic prompting to the regular Indic model (AI-Indic vs. AI-Indic-Demographic), the difference is not significant at all.}


\boldpara{Takeaway}
\chadded{Taken together, these analyses show that even in a downstream co-writing setting, Indic models do not preserve Indian writing styles any better than a global model; their suggestions often reproduce a Westernized gaze on Indian culture. Consistent with our broader results, prompting strategies provide surface-level alignment (e.g., more Indian artifacts) but leave deeper, entrenched cultural biases unchanged.}

\section{Discussion}
Our central question was whether region-specific training produces language models that are \emph{culturally} and not merely \emph{linguistically} aligned with their target communities. chadded{Across four complementary population-scale benchmarks and a downstream co-writing experiment}, we find that the answer is consistently negative. Indic models fluently \emph{speak} Indian languages, yet they reason, express opinions, recall cultural facts\chadded{, and even assist Indian writers} in ways that resemble Western-centric models. For high-stakes domains, this gap is not cosmetic; culturally misaligned models can misinform users, erode trust, and reproduce epistemic harms under the guise of localization. For the AI community, this underscores the need to treat culture as a first-class design target, on par with other tasks and metrics. We now discuss implications for human-centered AI evaluation and design.

\boldpara{Human-grounded AI evaluation at population scale}
Recent HCI work argues many AI evaluations of culture are ``thin'', i.e., they are detached from the community, and calls for \emph{thick} evaluations co-constructed with local participants~\cite{qadri2025thick}. We expand this idea into a thick $\times$ \emph{wide} evaluation blueprint. In particular, much of HCI's evaluation of AI has relied on small lab or field studies that richly capture context but are hard to scale across cultures and regions. We complement that depth with scale by grounding model evaluations in nationally representative surveys such as the World Values Survey and Pew Global Attitudes Survey~\cite{wvs, PewResearch2025Methodology}. These instruments are designed and fielded by on-the-ground teams and capture cultural values, beliefs, and practices from thousands of participants across dozens of countries, enabling us to anchor model outputs to population-level distributions rather than isolated individual judgments.
This approach does not replace qualitative or participatory inquiry but extends it, allowing HCI to connect lived experience with population-scale patterns.
We argue that pairing community workshops and participatory evaluations (thick) with survey-grounded, nationally representative instruments (wide) can truly center the lived experiences of local participants while evaluating AI models at scale.

Further, value-aligned models face the challenge of pluralism: a plurality of values and practices exists in the world, as well as within a nation-state. Sometimes, even conflicting cultural norms coexist within the same geography. This raises the question, who should models represent? \citet{sorensen2024roadmap} argued that models can either reflect the plurality of views in every output (overton pluralism), be steerable to a user's desired views (steerable pluralism), or reflect the plurality of views proportional to the population's preferences (distributional pluralism). Human-grounded data at the population scale is crucial to covering these diverse perspectives and building AI that is culturally situated, trustworthy, and socially responsive~\cite{irani2010postcolonial, dourish2016algorithms, friedman_vsd}.


\boldpara{Why evaluate ``small'' regional models at all?}
One might argue that current regional LLMs are too small (7–24B parameters) and far less capable than frontier models to be relevant. Why, then, should we evaluate these models? However, two factors make this evaluation \chreplaced{especially important}{not just necessary, but urgent}. First, these models are promoted as culturally grounded and tailored for regional users, yet, in practice, they are evaluated primarily on linguistic tasks. Our results show empirically that regional language fluency \(\neq\) cultural competence, and we hope this work encourages developers to treat cultural evaluation as a first-class concern \emph{during} model building. Model evaluation must therefore expand beyond task metrics and test cultural performance on thick (communally-grounded) and wide (population-scale) human data. Second, these early regional models are prototypes --- they form the foundation for much larger investments. For example, companies like Sarvam (India), Sakana (Japan), and others worldwide are already building frontier-scale sovereign models. Before pouring resources into scaling, it is \chreplaced{important to assess}{imperative to ask} whether these prototypes have succeeded at their core promise. \chreplaced{This will help}{It is prudent to} detect cultural brittleness \emph{before} building frontier-scale versions which could fossilize the cultural misalignment.

\boldpara{Training data and sovereign models}
In our results, comparisons against Indic models' base checkpoints show that supervised fine-tuning (SFT) on local-language data leaves cultural knowledge unchanged or degraded (Table~\ref{tab:cultbench-base_indic_comparison}). Digging deeper into the technical reports for Indic models, we find that almost every Indic model relies on instruction-tuning data that is either translated from English or tiny compared to the Western-centric pre-training data they attempt to counterbalance. For example, \model{nanda-10b}, \model{airavata-7b}, and \model{gajendra-7b} machine-translate English instruction datasets into Hindi or other Indic languages, with limited \emph{native} Hindi corpora (e.g., 6k Hindi WikiHow articles). \model{aya-expanse-8b} includes a sizeable, genuinely multilingual corpus, yet it tilts US-centric in our evaluation. \model{krutrim-2-12b} seems to be pre-trained from scratch by an Indian company, yet remains strongly US-leaning.

Thus, we hypothesize that the bottleneck lies in the imbalance of pretraining and fine-tuning data. Web-scale corpora used for pre-training are overwhelmingly produced by Western internet users, and fine-tuning on a thin slice of translated regional languages does not offset that prior. This mirrors HCI's long-standing critique that artifacts embed the assumptions of their origin contexts~\cite{dourish2016algorithms}. \chreplaced{Our findings reinforce}{We caution model developers} that building ``sovereign'' models is guided by Richard Sutton's \emph{bitter lesson}~\cite{Sutton2019Bitter}: more data usually beats clever algorithms. Hence, \chreplaced{improved cultural alignment likely depends on access to}{it is important to invest in} high-quality data from local populations that reflects their culture, values, and lived realities.

\boldpara{Implications for HCI and NLP}
\chadded{Practitioners can embed ``thick×wide'' cultural alignment into everyday workflows. For example, AI developers can evaluate any new models with a small participatory cohort (thick) and on a population-grounded survey dataset (wide) to track cultural drift before deployment.
Wide evaluations can show whether models reflect population-level cultural patterns, which values and norms they reproduce, and where systematic misalignments appear. Thick evaluations, informed by postcolonial critiques that challenge static or monolithic views of culture, can use community-engaged methods to surface subcultural variation and to assess users' perceptions of cultural (mis)alignment, including how these misalignments shape end-user tasks done with AI assistance. Together, 
this combined view can help identify when model generations feel culturally off and how such misalignments affect real-world use. Widely available datasets such as WVS, Pew, regional opinion polls, and census data already provide strong population-level priors, and partnerships with public institutions and civil-society organizations can help create more representative models as AI systems become embedded in high-stakes non-Western contexts~\cite{dennison2025designing}.}


\chadded{A complementary direction extends the thick×wide approach from evaluation to ownership by building community-centered auditing infrastructures. This aligns with value-sensitive design and postcolonial computing, which argue that communities should not only be sites of data collection but should hold authority over how cultural knowledge is defined, interpreted, and used. 
The purposes of these infrastructures is not free labor for AI companies but to shift power by giving communities mechanisms to detect misrepresentation, negotiate boundaries on data use, and refuse forms of cultural extraction. Community stewardship models already exist in adjacent domains, including local linguistic archives, community data trusts, and participatory knowledge repositories (e.g., AImpower). These models demonstrate that communities can maintain data resources that serve their own interests rather than those of commercial actors.}

\chadded{Building on these precedents, local NGOs, universities, and civil-society organizations can steward small, regularly updated corpora of culturally grounded prompts, examples, and norms. Such stewardship allows communities to decide what cultural knowledge is recorded, how it is governed, and under what conditions external actors may access it. Such community-centered auditing infrastructures can protect communities from cultural flattening and misuse.}

\boldpara{Limitations and Future Work}
Our analysis is constrained by the availability of regional models. There exist a few frontier-scale LLMs that were trained outside the West (e.g., DeepSeek), but no such model exists yet for India that we could add to our evaluation set. Cultural benchmarks like CulturalBench and NormAd used in our study, though community-grounded, are limited in size for many communities. While we mitigated this problem by triangulating across multiple experiments, we welcome future work that can create more of these ``thick'' benchmarks at scale (``wide'').\chadded{ Yet, we inherit the limitations of our datasets---their coarse granularity and tendency to treat countries as singular cultural units. We encourage work in cultural HCI and NLP that develops more fine-grained, subnational approaches to cultural analysis.} \chadded{ Finally, our study compares one non-Western nation (India) with one Western nation (the United States) due to the limited availability of non-Western models and benchmark datasets; we encourage similar evaluations for other regions as more regional models become available. }

\section{Conclusion}


Regional language fluency does not imply cultural competence. Across four human-grounded evaluations (leveraging nationally representative surveys and culturally verified QA) and a human participant study, we find that six Indic LLMs do not align better with Indian \emph{values} or \emph{practices} than global baselines, and sometimes perform worse. Prompting offered little relief, highlighting the limits of small-scale or translated local data in overcoming the Western bias baked into web-scale pretraining. Our findings position cultural alignment not as an add-on but as a core goal in model development. Building truly sovereign LLMs will require \emph{untranslated} regional corpora, training objectives that prioritize cultural alignment, and \emph{thick $\times$ wide} evaluation protocols that pair community co-design with population-scale data. These efforts align with HCI's commitment to culturally situated design. Only by embedding such practices early and continuously can future regional and sovereign models truly serve the people they are built for---not just linguistically, but also culturally and socially.


\bibliographystyle{ACM-Reference-Format}
\bibliography{refs}

@article{CultureLLM,
  title={Culturellm: Incorporating cultural differences into large language models},
  author={Li, Cheng and Chen, Mengzhuo and Wang, Jindong and Sitaram, Sunayana and Xie, Xing},
  journal={Advances in Neural Information Processing Systems},
  volume={37},
  pages={84799--84838},
  year={2024}
}

@inproceedings{cultural-alignment,
    title = "Cultural Alignment in Large Language Models: An Explanatory Analysis Based on Hofstede`s Cultural Dimensions",
    author = "Masoud, Reem  and
      Liu, Ziquan  and
      Ferianc, Martin  and
      Treleaven, Philip C.  and
      Rodrigues, Miguel Rodrigues",
    editor = "Rambow, Owen  and
      Wanner, Leo  and
      Apidianaki, Marianna  and
      Al-Khalifa, Hend  and
      Eugenio, Barbara Di  and
      Schockaert, Steven",
    booktitle = "Proceedings of the 31st International Conference on Computational Linguistics",
    month = jan,
    year = "2025",
    address = "Abu Dhabi, UAE",
    publisher = "Association for Computational Linguistics",
    url = "https://aclanthology.org/2025.coling-main.567/",
    pages = "8474--8503"
}

@misc{hofstede_wvs_llm,
      title={Probing Pre-Trained Language Models for Cross-Cultural Differences in Values}, 
      author={Arnav Arora and Lucie-Aimée Kaffee and Isabelle Augenstein},
      year={2023},
      eprint={2203.13722},
      archivePrefix={arXiv},
      primaryClass={cs.CL},
      url={https://arxiv.org/abs/2203.13722}, 
}

@inproceedings{ahuja-etal-2023-mega,
    title = "{MEGA}: Multilingual Evaluation of Generative {AI}",
    author = "Ahuja, Kabir  and
      Diddee, Harshita  and
      Hada, Rishav  and
      Ochieng, Millicent  and
      Ramesh, Krithika  and
      Jain, Prachi  and
      Nambi, Akshay  and
      Ganu, Tanuja  and
      Segal, Sameer  and
      Ahmed, Mohamed  and
      Bali, Kalika  and
      Sitaram, Sunayana",
    editor = "Bouamor, Houda  and
      Pino, Juan  and
      Bali, Kalika",
    booktitle = "Proceedings of the 2023 Conference on Empirical Methods in Natural Language Processing",
    month = dec,
    year = "2023",
    address = "Singapore",
    publisher = "Association for Computational Linguistics",
    url = "https://aclanthology.org/2023.emnlp-main.258/",
    doi = "10.18653/v1/2023.emnlp-main.258",
    pages = "4232--4267",
}

@misc{havaldar2023,
      title={Multilingual Language Models are not Multicultural: A Case Study in Emotion}, 
      author={Shreya Havaldar and Sunny Rai and Bhumika Singhal and Langchen Liu and Sharath Chandra Guntuku and Lyle Ungar},
      year={2023},
      eprint={2307.01370},
      archivePrefix={arXiv},
      primaryClass={cs.CL},
      url={https://arxiv.org/abs/2307.01370}, 
}

@misc{naous2024,
      title={Having Beer after Prayer? Measuring Cultural Bias in Large Language Models}, 
      author={Tarek Naous and Michael J. Ryan and Alan Ritter and Wei Xu},
      year={2024},
      eprint={2305.14456},
      archivePrefix={arXiv},
      primaryClass={cs.CL},
      url={https://arxiv.org/abs/2305.14456}, 
}

@misc{nanda2025,
      title={Llama-3-Nanda-10B-Chat: An Open Generative Large Language Model for Hindi}, 
      author={Monojit Choudhury and Shivam Chauhan and Rocktim Jyoti Das and Dhruv Sahnan and Xudong Han and Haonan Li and Aaryamonvikram Singh and Alok Anil Jadhav and Utkarsh Agarwal and Mukund Choudhary and Debopriyo Banerjee and Fajri Koto and Junaid Bhat and Awantika Shukla and Samujjwal Ghosh and Samta Kamboj and Onkar Pandit and Lalit Pradhan and Rahul Pal and Sunil Sahu and Soundar Doraiswamy and Parvez Mullah and Ali El Filali and Neha Sengupta and Gokul Ramakrishnan and Rituraj Joshi and Gurpreet Gosal and Avraham Sheinin and Natalia Vassilieva and Preslav Nakov},
      year={2025},
      eprint={2504.06011},
      archivePrefix={arXiv},
      primaryClass={cs.CL},
      url={https://arxiv.org/abs/2504.06011}, 
}

@article{Tao2024,
  title = {Cultural bias and cultural alignment of large language models},
  volume = {3},
  ISSN = {2752-6542},
  url = {http://dx.doi.org/10.1093/pnasnexus/pgae346},
  DOI = {10.1093/pnasnexus/pgae346},
  number = {9},
  journal = {PNAS Nexus},
  publisher = {Oxford University Press (OUP)},
  author = {Tao,  Yan and Viberg,  Olga and Baker,  Ryan S and Kizilcec,  René F},
  editor = {Muthukrishna,  Michael},
  year = {2024},
  month = sep 
}

@misc{wvs,
  doi = {10.14281/18241.23},
  url = {https://www.worldvaluessurvey.org/WVSEVStrend.jsp},
  author = {Haerpfer,  Christian and Inglehart,  Ronald and Moreno,  Alejandro and Welzel,  Christian and Kizilova,  Kseniya and Diez-Medrano,  Jaime and Lagos,  Marta and Norris,  Pippa and Ponarin,  Eduard and Puranen,  Bi},
  language = {en},
  title = {World Values Survey (1981-2022). Trend File},
  publisher = {World Values Survey Association},
  year = {2022},
  copyright = {Other}
}

@book{Inglehart2005,
  title = {Modernization,  Cultural Change,  and Democracy: The Human Development Sequence},
  ISBN = {9780511790881},
  url = {http://dx.doi.org/10.1017/CBO9780511790881},
  DOI = {10.1017/cbo9780511790881},
  publisher = {Cambridge University Press},
  author = {Inglehart,  Ronald and Welzel,  Christian},
  year = {2005},
  month = jan 
}

@misc{Durmus2023,
  doi = {10.48550/arxiv.2306.16388},
  url = {https://arxiv.org/abs/2306.16388},
  author = {Durmus,  Esin and Nguyen,  Karina and Liao,  Thomas I. and Schiefer,  Nicholas and Askell,  Amanda and Bakhtin,  Anton and Chen,  Carol and Hatfield-Dodds,  Zac and Hernandez,  Danny and Joseph,  Nicholas and Lovitt,  Liane and McCandlish,  Sam and Sikder,  Orowa and Tamkin,  Alex and Thamkul,  Janel and Kaplan,  Jared and Clark,  Jack and Ganguli,  Deep},
  title = {Towards Measuring the Representation of Subjective Global Opinions in Language Models},
  publisher = {arXiv},
  year = {2023},
  copyright = {Creative Commons Attribution 4.0 International}
}

@misc{CulturalBench,
  doi = {10.48550/arxiv.2410.02677},
  url = {https://arxiv.org/abs/2410.02677},
  author = {Chiu,  Yu Ying and Jiang,  Liwei and Lin,  Bill Yuchen and Park,  Chan Young and Li,  Shuyue Stella and Ravi,  Sahithya and Bhatia,  Mehar and Antoniak,  Maria and Tsvetkov,  Yulia and Shwartz,  Vered and Choi,  Yejin},
  title = {CulturalBench: a Robust,  Diverse and Challenging Benchmark on Measuring the (Lack of) Cultural Knowledge of LLMs},
  publisher = {arXiv},
  year = {2024},
  copyright = {Creative Commons Attribution 4.0 International}
}

@misc{NormAd,
  doi = {10.48550/arxiv.2404.12464},
  url = {https://arxiv.org/abs/2404.12464},
  author = {Rao,  Abhinav and Yerukola,  Akhila and Shah,  Vishwa and Reinecke,  Katharina and Sap,  Maarten},
  title = {NormAd: A Framework for Measuring the Cultural Adaptability of Large Language Models},
  publisher = {arXiv},
  year = {2024},
  copyright = {Creative Commons Attribution 4.0 International}
}

@misc{cao2023,
      title={Assessing Cross-Cultural Alignment between ChatGPT and Human Societies: An Empirical Study}, 
      author={Yong Cao and Li Zhou and Seolhwa Lee and Laura Cabello and Min Chen and Daniel Hershcovich},
      year={2023},
      eprint={2303.17466},
      archivePrefix={arXiv},
      primaryClass={cs.CL},
      url={https://arxiv.org/abs/2303.17466}, 
}

@misc{johnson2022,
      title={The Ghost in the Machine has an American accent: value conflict in GPT-3}, 
      author={Rebecca L Johnson and Giada Pistilli and Natalia Menédez-González and Leslye Denisse Dias Duran and Enrico Panai and Julija Kalpokiene and Donald Jay Bertulfo},
      year={2022},
      eprint={2203.07785},
      archivePrefix={arXiv},
      primaryClass={cs.CL},
      url={https://arxiv.org/abs/2203.07785}, 
}

@inproceedings{Qadri_2023, series={FAccT ’23},
   title={AI’s Regimes of Representation: A Community-centered Study of Text-to-Image Models in South Asia},
   url={http://dx.doi.org/10.1145/3593013.3594016},
   DOI={10.1145/3593013.3594016},
   booktitle={2023 ACM Conference on Fairness, Accountability, and Transparency},
   publisher={ACM},
   author={Qadri, Rida and Shelby, Renee and Bennett, Cynthia L. and Denton, Emily},
   year={2023},
   month=jun, pages={506–517},
   collection={FAccT ’23} }

@inproceedings{Agarwal_2025, series={CHI ’25},
   title={{AI Suggestions Homogenize Writing Toward Western Styles and Diminish Cultural Nuances}},
   url={http://dx.doi.org/10.1145/3706598.3713564},
   DOI={10.1145/3706598.3713564},
   booktitle={Proceedings of the 2025 CHI Conference on Human Factors in Computing Systems},
   publisher={ACM},
   author={Agarwal, Dhruv and Naaman, Mor and Vashistha, Aditya},
   year={2025},
   month=apr, pages={1–21},
   collection={CHI ’25} }

@misc{jais,
      title={Jais and Jais-chat: Arabic-Centric Foundation and Instruction-Tuned Open Generative Large Language Models}, 
      author={Neha Sengupta and Sunil Kumar Sahu and Bokang Jia and Satheesh Katipomu and Haonan Li and Fajri Koto and William Marshall and Gurpreet Gosal and Cynthia Liu and Zhiming Chen and Osama Mohammed Afzal and Samta Kamboj and Onkar Pandit and Rahul Pal and Lalit Pradhan and Zain Muhammad Mujahid and Massa Baali and Xudong Han and Sondos Mahmoud Bsharat and Alham Fikri Aji and Zhiqiang Shen and Zhengzhong Liu and Natalia Vassilieva and Joel Hestness and Andy Hock and Andrew Feldman and Jonathan Lee and Andrew Jackson and Hector Xuguang Ren and Preslav Nakov and Timothy Baldwin and Eric Xing},
      year={2023},
      eprint={2308.16149},
      archivePrefix={arXiv},
      primaryClass={cs.CL},
      url={https://arxiv.org/abs/2308.16149}, 
}

@book{Hofstede:1991,
  address = {London and New York},
  author = {Hofstede, Geert H.},
  booktitle = {Cultures and Organizations: Software of the Mind},
  pages = {xii, 279},
  publisher = {McGraw-Hill},
  title = {Cultures and organizations: Software of the mind},
  year = 1991
}

@misc{mukherjee2024placebo,
      title={Cultural Conditioning or Placebo? On the Effectiveness of Socio-Demographic Prompting}, 
      author={Sagnik Mukherjee and Muhammad Farid Adilazuarda and Sunayana Sitaram and Kalika Bali and Alham Fikri Aji and Monojit Choudhury},
      year={2024},
      eprint={2406.11661},
      archivePrefix={arXiv},
      primaryClass={cs.CL},
      url={https://arxiv.org/abs/2406.11661}, 
}

@misc{adilazuarda2024survey,
      title={Towards Measuring and Modeling "Culture" in LLMs: A Survey}, 
      author={Muhammad Farid Adilazuarda and Sagnik Mukherjee and Pradhyumna Lavania and Siddhant Singh and Alham Fikri Aji and Jacki O'Neill and Ashutosh Modi and Monojit Choudhury},
      year={2024},
      eprint={2403.15412},
      archivePrefix={arXiv},
      primaryClass={cs.CY},
      url={https://arxiv.org/abs/2403.15412}, 
}

@inproceedings{saha-etal-2025-meta,
    title = "Meta-Cultural Competence: Climbing the Right Hill of Cultural Awareness",
    author = "Saha, Sougata  and
      Pandey, Saurabh Kumar  and
      Choudhury, Monojit",
    editor = "Chiruzzo, Luis  and
      Ritter, Alan  and
      Wang, Lu",
    booktitle = "Proceedings of the 2025 Conference of the Nations of the Americas Chapter of the Association for Computational Linguistics: Human Language Technologies (Volume 1: Long Papers)",
    month = apr,
    year = "2025",
    address = "Albuquerque, New Mexico",
    publisher = "Association for Computational Linguistics",
    url = "https://aclanthology.org/2025.naacl-long.408/",
    pages = "8025--8042",
    ISBN = "979-8-89176-189-6"
}

@inproceedings{Hershcovich2022Challenges,
    title = "Challenges and Strategies in Cross-Cultural {NLP}",
    author = "Hershcovich, Daniel  and
      Frank, Stella  and
      Lent, Heather  and
      de Lhoneux, Miryam  and
      Abdou, Mostafa  and
      Brandl, Stephanie  and
      Bugliarello, Emanuele  and
      Cabello Piqueras, Laura  and
      Chalkidis, Ilias  and
      Cui, Ruixiang  and
      Fierro, Constanza  and
      Margatina, Katerina  and
      Rust, Phillip  and
      S{\o}gaard, Anders",
    editor = "Muresan, Smaranda  and
      Nakov, Preslav  and
      Villavicencio, Aline",
    booktitle = "Proceedings of the 60th Annual Meeting of the Association for Computational Linguistics (Volume 1: Long Papers)",
    month = may,
    year = "2022",
    address = "Dublin, Ireland",
    publisher = "Association for Computational Linguistics",
    url = "https://aclanthology.org/2022.acl-long.482/",
    doi = "10.18653/v1/2022.acl-long.482",
    pages = "6997--7013"
}

@misc{beyond_metrics_culturally_nuanced,
      title={Beyond Metrics: Evaluating LLMs' Effectiveness in Culturally Nuanced, Low-Resource Real-World Scenarios}, 
      author={Millicent Ochieng and Varun Gumma and Sunayana Sitaram and Jindong Wang and Vishrav Chaudhary and Keshet Ronen and Kalika Bali and Jacki O'Neill},
      year={2024},
      eprint={2406.00343},
      archivePrefix={arXiv},
      primaryClass={cs.CL},
      url={https://arxiv.org/abs/2406.00343}, 
}

@inproceedings{liu2024understanding,
  title={Understanding public perceptions of AI conversational agents: A cross-cultural analysis},
  author={Liu, Zihan and Li, Han and Chen, Anfan and Zhang, Renwen and Lee, Yi-Chieh},
  booktitle={Proceedings of the 2024 CHI conference on human factors in computing systems},
  pages={1--17},
  year={2024}
}

@article{dourish2016algorithms,
  title={Algorithms and their others: Algorithmic culture in context},
  author={Dourish, Paul},
  journal={Big Data \& Society},
  volume={3},
  number={2},
  pages={2053951716665128},
  year={2016},
  publisher={SAGE Publications Sage UK: London, England}
}

@inproceedings{irani2010postcolonial,
author = {Irani, Lilly and Vertesi, Janet and Dourish, Paul and Philip, Kavita and Grinter, Rebecca E.},
title = {Postcolonial computing: a lens on design and development},
year = {2010},
isbn = {9781605589299},
publisher = {Association for Computing Machinery},
address = {New York, NY, USA},
url = {https://doi.org/10.1145/1753326.1753522},
doi = {10.1145/1753326.1753522},
booktitle = {Proceedings of the SIGCHI Conference on Human Factors in Computing Systems},
pages = {1311–1320},
numpages = {10},
location = {Atlanta, Georgia, USA},
series = {CHI '10}
}

@misc{sambasivan2021reimagining,
      title={Re-imagining Algorithmic Fairness in India and Beyond}, 
      author={Nithya Sambasivan and Erin Arnesen and Ben Hutchinson and Tulsee Doshi and Vinodkumar Prabhakaran},
      year={2021},
      eprint={2101.09995},
      archivePrefix={arXiv},
      primaryClass={cs.CY},
      url={https://arxiv.org/abs/2101.09995}, 
}

@article{Marcus2000,
  title = {Crosscurrents: cultural dimensions and global Web user-interface design},
  volume = {7},
  ISSN = {1558-3449},
  url = {http://dx.doi.org/10.1145/345190.345238},
  DOI = {10.1145/345190.345238},
  number = {4},
  journal = {Interactions},
  publisher = {Association for Computing Machinery (ACM)},
  author = {Marcus,  Aaron and Gould,  Emilie West},
  year = {2000},
  month = jul,
  pages = {32–46}
}

@article{Clemmensen2009,
  title = {Cultural cognition in usability evaluation},
  volume = {21},
  ISSN = {0953-5438},
  url = {http://dx.doi.org/10.1016/j.intcom.2009.05.003},
  DOI = {10.1016/j.intcom.2009.05.003},
  number = {3},
  journal = {Interacting with Computers},
  publisher = {Oxford University Press (OUP)},
  author = {Clemmensen,  Torkil and Hertzum,  Morten and Hornbæk,  Kasper and Shi,  Qingxin and Yammiyavar,  Pradeep},
  year = {2009},
  month = jul,
  pages = {212–220}
}

@book{Sun2012,
  title = {Cross-Cultural Technology DesignCreating Culture-Sensitive Technology for Local Users},
  ISBN = {9780199744763},
  url = {http://dx.doi.org/10.1093/acprof:oso/9780199744763.001.0001},
  DOI = {10.1093/acprof:oso/9780199744763.001.0001},
  publisher = {Oxford University Press},
  author = {Sun,  Huatong},
  year = {2012},
  month = mar 
}

@inproceedings{Qadri2025,
  series = {CHI ’25},
  title = {AI and Non-Western Art Worlds: Reimagining Critical AI Futures through Artistic Inquiry and Situated Dialogue},
  url = {http://dx.doi.org/10.1145/3706598.3714049},
  DOI = {10.1145/3706598.3714049},
  booktitle = {Proceedings of the 2025 CHI Conference on Human Factors in Computing Systems},
  publisher = {ACM},
  author = {Qadri,  Rida and Mirowski,  Piotr and Denton,  Remi},
  year = {2025},
  month = apr,
  pages = {1–17},
  collection = {CHI ’25}
}

@misc{saha2025reading,
      title={Reading between the Lines: Can LLMs Identify Cross-Cultural Communication Gaps?}, 
      author={Sougata Saha and Saurabh Kumar Pandey and Harshit Gupta and Monojit Choudhury},
      year={2025},
      eprint={2502.09636},
      archivePrefix={arXiv},
      primaryClass={cs.CL},
      url={https://arxiv.org/abs/2502.09636}, 
}

@misc{shen2025valueactiongap,
      title={Mind the Value-Action Gap: Do LLMs Act in Alignment with Their Values?}, 
      author={Hua Shen and Nicholas Clark and Tanushree Mitra},
      year={2025},
      eprint={2501.15463},
      archivePrefix={arXiv},
      primaryClass={cs.HC},
      url={https://arxiv.org/abs/2501.15463}, 
}

@inproceedings{Khan2025,
  series = {FAccT ’25},
  title = {Randomness,  Not Representation: The Unreliability of Evaluating Cultural Alignment in LLMs},
  url = {http://dx.doi.org/10.1145/3715275.3732147},
  DOI = {10.1145/3715275.3732147},
  booktitle = {Proceedings of the 2025 ACM Conference on Fairness,  Accountability,  and Transparency},
  publisher = {ACM},
  author = {Khan,  Ariba and Casper,  Stephen and Hadfield-Menell,  Dylan},
  year = {2025},
  month = jun,
  pages = {2151–2165},
  collection = {FAccT ’25}
}

@misc{qadri2025thick,
      title={The Case for "Thick Evaluations" of Cultural Representation in AI}, 
      author={Rida Qadri and Mark Diaz and Ding Wang and Michael Madaio},
      year={2025},
      eprint={2503.19075},
      archivePrefix={arXiv},
      primaryClass={cs.CY},
      url={https://arxiv.org/abs/2503.19075}, 
}

@inproceedings{Song2025Good,
    title = "The Good, The Bad, and The Greedy: Evaluation of {LLM}s Should Not Ignore Non-Determinism",
    author = "Song, Yifan  and
      Wang, Guoyin  and
      Li, Sujian  and
      Lin, Bill Yuchen",
    editor = "Chiruzzo, Luis  and
      Ritter, Alan  and
      Wang, Lu",
    booktitle = "Proceedings of the 2025 Conference of the Nations of the Americas Chapter of the Association for Computational Linguistics: Human Language Technologies (Volume 1: Long Papers)",
    month = apr,
    year = "2025",
    address = "Albuquerque, New Mexico",
    publisher = "Association for Computational Linguistics",
    url = "https://aclanthology.org/2025.naacl-long.211/",
    doi = "10.18653/v1/2025.naacl-long.211",
    pages = "4195--4206",
    ISBN = "979-8-89176-189-6"
}

@misc{PewResearch2025Methodology,
  author = "{Pew Research Center}",
  title = "{Methodology: Spring 2025 Global Attitudes Survey}",
  howpublished = "\url{https://www.pewresearch.org/2025/08/19/methodology-global-threats-2025/}",
  note = "Accessed: 2025-09-11",
  year = "2025"
}

@misc{Sutton2019Bitter,
  author = {Sutton, Richard},
  title = {The Bitter Lesson},
  year = {2019},
  howpublished = {\url{http://www.incompleteideas.net/IncIdeas/BitterLesson.html}},
  note = {Retrieved September 7, 2025}
}

@misc{sorensen2024roadmap,
      title={A Roadmap to Pluralistic Alignment}, 
      author={Taylor Sorensen and Jared Moore and Jillian Fisher and Mitchell Gordon and Niloofar Mireshghallah and Christopher Michael Rytting and Andre Ye and Liwei Jiang and Ximing Lu and Nouha Dziri and Tim Althoff and Yejin Choi},
      year={2024},
      eprint={2402.05070},
      archivePrefix={arXiv},
      primaryClass={cs.AI},
      url={https://arxiv.org/abs/2402.05070}, 
}

@BOOK{friedman_vsd,
  title     = "Value sensitive design",
  author    = "Friedman, Batya and Hendry, David G",
  publisher = "MIT Press",
  series    = "The MIT Press",
  month     =  may,
  year      =  2019,
  address   = "London, England",
  language  = "en"
}

@article{Beugelsdijk2018,
  title = {Dimensions and Dynamics of National Culture: Synthesizing Hofstede With Inglehart},
  volume = {49},
  ISSN = {1552-5422},
  url = {http://dx.doi.org/10.1177/0022022118798505},
  DOI = {10.1177/0022022118798505},
  number = {10},
  journal = {Journal of Cross-Cultural Psychology},
  publisher = {SAGE Publications},
  author = {Beugelsdijk,  Sjoerd and Welzel,  Chris},
  year = {2018},
  month = oct,
  pages = {1469–1505}
}

@misc{dennison2025designing,
      title={Designing Culturally Aligned AI Systems For Social Good in Non-Western Contexts}, 
      author={Deepak Varuvel Dennison and Mohit Jain and Tanuja Ganu and Aditya Vashistha},
      year={2025},
      eprint={2509.16158},
      archivePrefix={arXiv},
      primaryClass={cs.HC},
      url={https://arxiv.org/abs/2509.16158}, 
}

\appendix

\section{Cultural Map Experiment} \label{appendix:iw_cultural_map}

\begin{figure*}[t]
    \centering
    \begin{tabular}{cc}
    \includegraphics[width=0.4\textwidth]{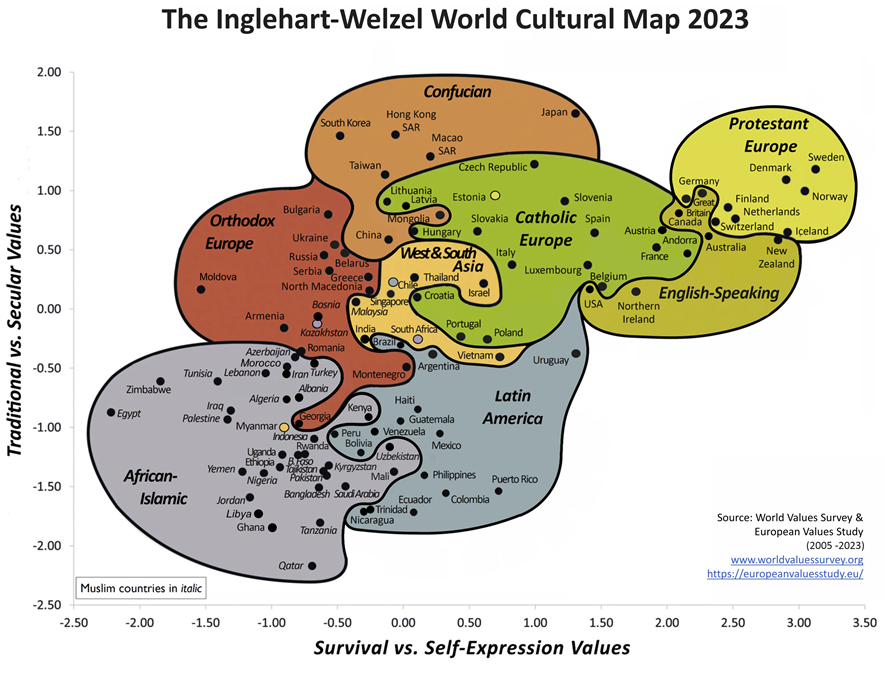} & \includegraphics[width=0.5\textwidth]{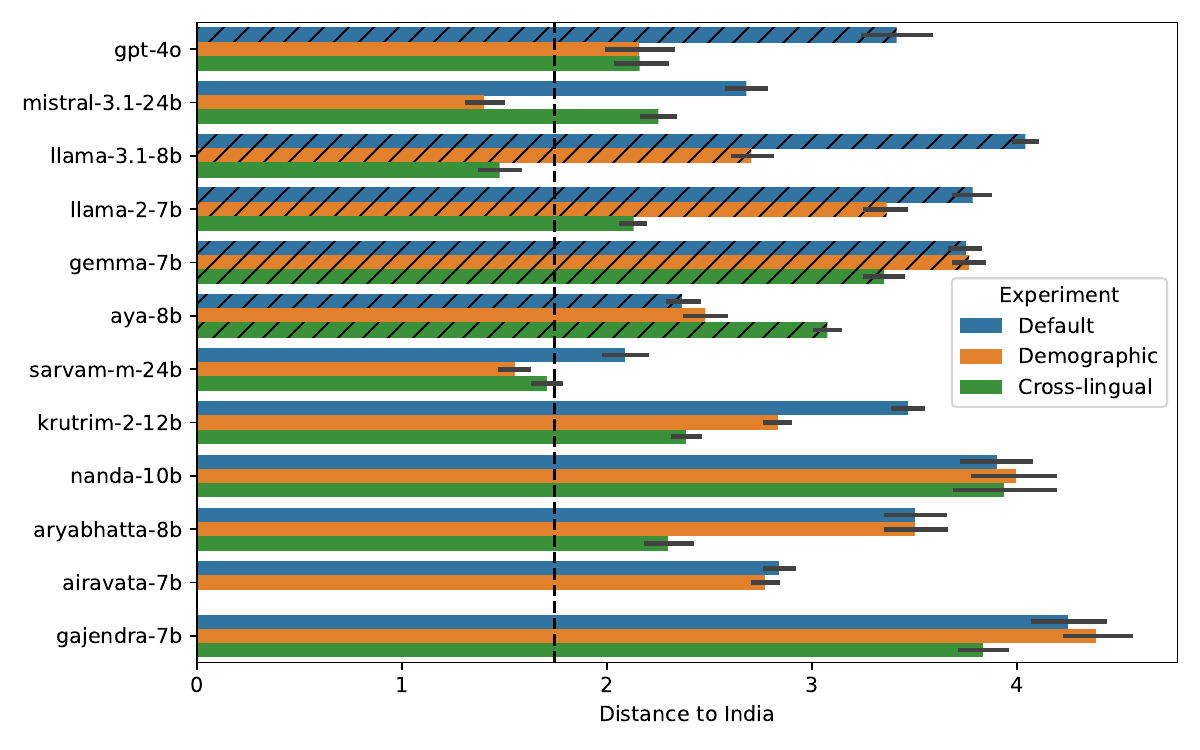}  \\
    (a) Inglehart-Welzel Cultural Map 2023~\cite{Inglehart2005} & (b) Distance from India across prompting strategies
    \end{tabular}
    \caption{(a) Inglehart-Welzel Cultural Map 2023~\cite{Inglehart2005}. (b) Euclidean distance between each model and India under three prompting strategies. Lower is better.\chadded{ The dotted vertical line marks the distance between India and US on the cultural map; most models are farther from India than an average American.}}
    \label{fig:cultural_map_india_distances}
\end{figure*}

\subsection{Inglehart-Welzel Cultural Map}
Refer to Figure~\ref{fig:cultural_map_india_distances}a for an example of the cultural map from 2023.

\subsection{Distance from India across Prompting Strategies} \label{appendix:culturalmap_india_distances}
Refer to Figure~\ref{fig:cultural_map_india_distances}b.

\subsection{Distance from India and the US} \label{appendix:culturalmap_distances}
\begin{figure*}[t]
    \centering
    \includegraphics[width=\linewidth]{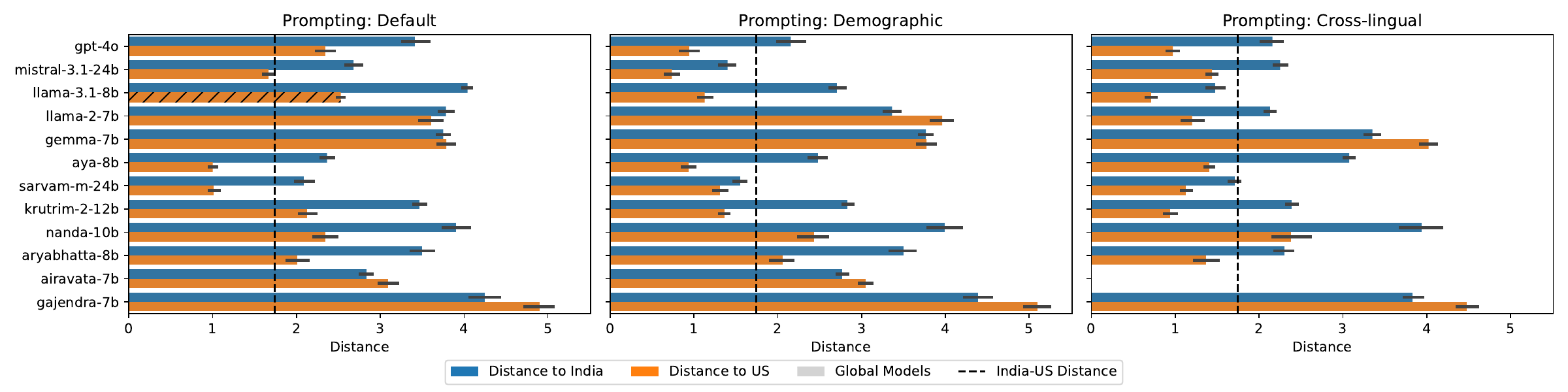}
    \caption{Distance of each model from India and the US across the three prompting strategies.\chadded{ The dotted vertical line marks the distance between India and US on the cultural map.}}
    \label{fig:culturalmap_distances}
\end{figure*}
Figure~\ref{fig:culturalmap_distances} shows the distance of each model from India and the US on the IW cultural map.

\section{Opinion Alignment Experiment} \label{appendix:opinion_alignment}




\subsection{Impact of Prompting Strategies} \label{appendix:nCAD_all_prompting_strategies}
\begin{figure*}[h]
    \centering
    \includegraphics[width=\textwidth]{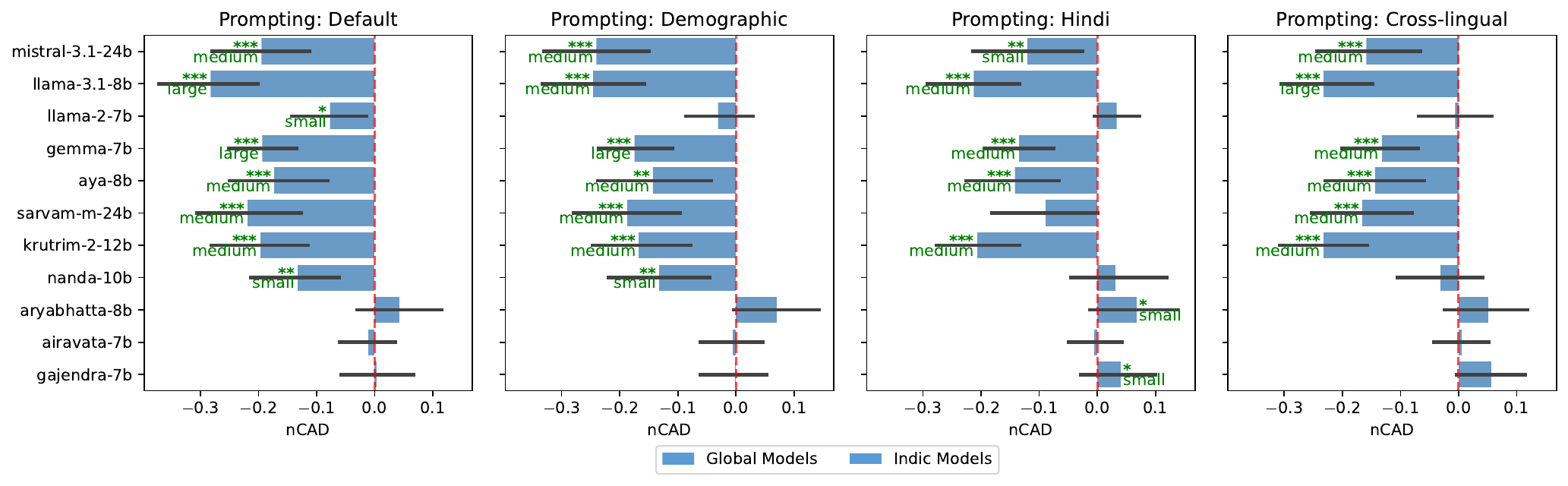}
    \caption{Normalized CAD (nCAD) under four prompting strategies. Negative values indicate a US tilt, and positive values indicate an India tilt.\chadded{ The dotted vertical line at $x=0$ represents no tilt.} * $p<0.05$, ** $p<0.01$, *** $p<0.001$ (two-sided). Tests used are the same as in Figure~\ref{fig:globalopinionqa}.}
    \label{fig:nCAD_all_prompting_strategies}
\end{figure*}
Refer to Figure~\ref{fig:nCAD_all_prompting_strategies}.

\subsection{Raw Similarity Scores} \label{appendix:globalopinionqa_appendix}
\begin{figure}[h]
    \centering
    \includegraphics[width=0.6\linewidth]{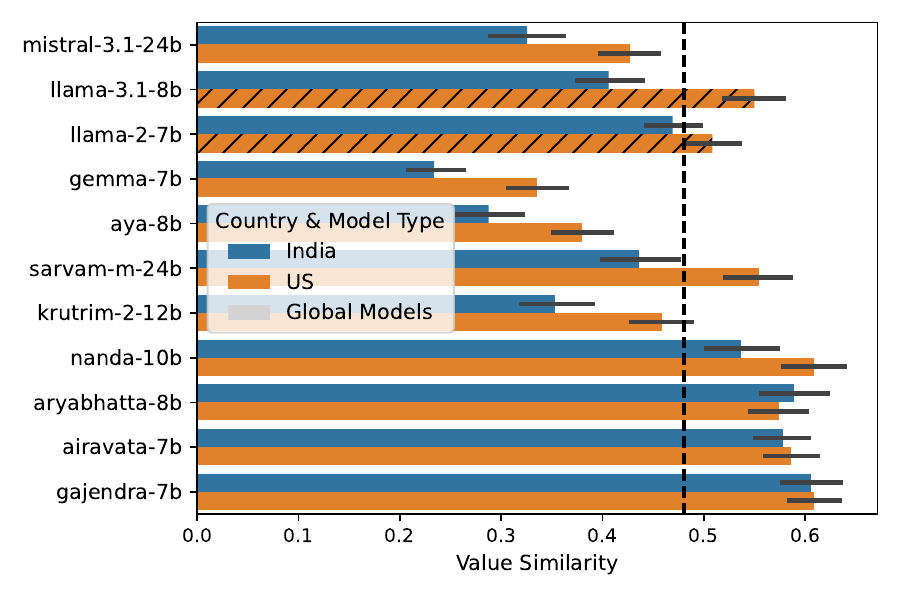}
    \caption{Raw similarity scores between each model's responses and Indian and American ground-truths (before computing CAD) under Default prompting. In other words, \(\operatorname{Sim}(m, ind)\) and \(\operatorname{Sim}(m, usa)\) for all models $m$. The dotted line represents the inter-country agreement between Indians and Americans (0.48).}
    \label{fig:globalopinionqa_appendix}
\end{figure}
\begin{table*}[t]
    \small
    \centering
    \begin{tabular}{l*{4}{p{2.4cm}}}
        \toprule
        Model & Default & Demographic & Hindi & Cross-lingual \\
        \midrule
        \model{mistral-3.1-24b} & 0.33±0.17 // 0.43±0.15 & 0.33±0.17 // 0.46±0.15 & 0.40±0.16 // 0.47±0.15 & 0.38±0.16 // 0.46±0.15 \\
        \model{llama-3.1-8b} & 0.41±0.17 // 0.55±0.15 & 0.46±0.17 // 0.58±0.14 & 0.53±0.17 // 0.64±0.15 & 0.50±0.16 // 0.62±0.15 \\
        \model{llama-2-7b} & 0.47±0.13 // 0.51±0.13 & 0.47±0.12 // 0.49±0.13 & 0.48±0.09 // 0.47±0.10 & 0.48±0.11 // 0.48±0.14 \\
        \model{gemma-7b} & 0.23±0.13 // 0.34±0.15 & 0.24±0.13 // 0.33±0.15 & 0.25±0.12 // 0.32±0.15 & 0.25±0.12 // 0.32±0.15 \\
        \model{aya-8b} & 0.29±0.16 // 0.38±0.15 & 0.30±0.17 // 0.38±0.16 & 0.27±0.14 // 0.34±0.16 & 0.29±0.15 // 0.37±0.16 \\
        \model{sarvam-m-24b} & 0.44±0.19 // 0.55±0.16 & 0.47±0.17 // 0.57±0.15 & 0.46±0.18 // 0.51±0.15 & 0.47±0.18 // 0.56±0.16 \\
        \model{krutrim-2-12b} & 0.35±0.17 // 0.46±0.15 & 0.37±0.16 // 0.46±0.15 & 0.36±0.15 // 0.47±0.15 & 0.36±0.16 // 0.48±0.15 \\
        \model{nanda-10b} & 0.54±0.17 // 0.61±0.15 & 0.54±0.17 // 0.61±0.15 & 0.59±0.18 // 0.58±0.15 & 0.57±0.17 // 0.59±0.14 \\
        \model{aryabhatta-8b} & 0.59±0.16 // 0.57±0.13 & 0.58±0.16 // 0.55±0.15 & 0.62±0.16 // 0.60±0.13 & 0.62±0.16 // 0.60±0.13 \\
        \model{airavata-7b} & 0.58±0.13 // 0.59±0.13 & 0.58±0.13 // 0.58±0.13 & 0.58±0.12 // 0.58±0.12 & 0.58±0.13 // 0.58±0.13 \\
        \model{gajendra-7b} & 0.61±0.14 // 0.61±0.12 & 0.60±0.13 // 0.61±0.12 & 0.58±0.14 // 0.56±0.13 & 0.63±0.14 // 0.60±0.12 \\
        \bottomrule
    \end{tabular}
    \caption{Mean and standard deviation of \(\operatorname{Sim}(m, ind)\) (first) and \(\operatorname{Sim}(m, usa)\) (second) scores for all models under each prompting strategy, prior to CAD normalization. Higher values indicate greater similarity to the corresponding population's opinions. Similarity with Indian opinion is generally lesser than American opinion, resulting in negative CAD scores in Figure~\ref{fig:globalopinionqa}.}
    \label{tab:opinion_raw_results}
\end{table*}

Inspecting the raw similarity scores in Figure~\ref{fig:globalopinionqa_appendix} (before computing the CAD), we find that some models (e.g. \model{gemma-7b}, \model{aya-8b}, \model{krutrim-2-12b}) are generally misaligned with both Indian and American responses. These models do not echo any of the target groups. These models exhibit lower average similarity than even the inter-country agreement between Indians and Americans (0.48). In other words, American opinions are a better proxy for Indian public opinion (and vice versa) than these models' predictions. This might not always be a bad thing. In fact, it might be desirable for a global model not to be aligned to any culture than to be strongly aligned to one specific culture. However, it constitutes a failure mode for regional models (e.g., Indic models), which are explicitly designed to align with a specific user group. The raw numbers in this plot (and for the other three prompting strategies) are shown in Table~\ref{tab:opinion_raw_results}.

\section{Cultural Knowledge Experiment} \label{appendix:normad}

\begin{table*}[t]
    \small
    \centering
    \begin{tabular}{llll}
    \toprule
    Model & Default & Demographic & Cross-lingual \\
    \midrule
    \model{gpt-4o} & 0.85±0.36 // 0.98±0.15 & 0.85±0.36 // 0.98±0.13 & 0.64±0.48 // 0.94±0.23 \\
    \model{mistral-3.1-24b} & 0.81±0.39 // 0.92±0.27 & 0.82±0.38 // 0.90±0.29 & 0.58±0.49 // 0.70±0.46 \\
    \model{llama-3.1-8b} & 0.73±0.45 // 0.86±0.34 & 0.76±0.43 // 0.88±0.32 & 0.58±0.49 // 0.63±0.48 \\
    \model{llama-2-7b} & 0.57±0.49 // 0.69±0.46 & 0.59±0.49 // 0.68±0.47 & 0.27±0.44 // 0.32±0.47 \\
    \model{gemma-7b} & 0.50±0.50 // 0.82±0.38 & 0.51±0.50 // 0.80±0.40 & 0.40±0.49 // 0.53±0.50 \\
    \model{aya-8b} & 0.68±0.47 // 0.91±0.28 & 0.68±0.47 // 0.90±0.29 & 0.50±0.50 // 0.74±0.44 \\
    \model{sarvam-m-24b} & 0.79±0.41 // 0.87±0.33 & 0.80±0.40 // 0.87±0.34 & 0.55±0.50 // 0.75±0.43 \\
    \model{krutrim-2-12b} & 0.73±0.45 // 0.84±0.36 & 0.73±0.44 // 0.83±0.37 & 0.56±0.50 // 0.73±0.44 \\
    \model{nanda-10b} & 0.61±0.49 // 0.76±0.43 & 0.63±0.48 // 0.77±0.42 & 0.54±0.50 // 0.68±0.46 \\
    \model{aryabhatta-8b} & 0.53±0.50 // 0.74±0.44 & 0.54±0.50 // 0.73±0.44 & 0.49±0.50 // 0.65±0.48 \\
    \model{airavata-7b} & 0.52±0.50 // 0.70±0.46 & 0.52±0.50 // 0.69±0.46 & 0.39±0.49 // 0.48±0.50 \\
    \model{gajendra-7b} & 0.45±0.50 // 0.59±0.49 & 0.45±0.50 // 0.57±0.49 & 0.29±0.45 // 0.31±0.46 \\
    \bottomrule
    \end{tabular}
    \caption{Accuracy (mean and standard deviation) on CulturalBench under all prompting strategies. Entries show accuracy on India (left) and US (right) specific questions.}
    \label{tab:culturalbench_raw_results}
\end{table*}

Accuracy results for the CulturalBench dataset shown in Figure~\ref{fig:culturalbench} are reported in Table~\ref{tab:culturalbench_raw_results}.

\section{Cultural Adaptation Experiment} \label{appendix:normad}
\begin{table*}[t]
    \small
    \centering
    \begin{tabular}{llll}
    \toprule
    Model & Country & Value & Rule-of-Thumb \\
    \midrule
    \model{gpt-4o} & 0.53±0.50 // 0.69±0.46 & 0.61±0.49 // 0.75±0.44 & 0.78±0.41 // 0.81±0.40 \\
    \model{mistral-3.1-24b} & 0.61±0.49 // 0.56±0.50 & 0.70±0.46 // 0.79±0.41 & 0.78±0.41 // 0.88±0.32 \\
    \model{llama-3.1-8b} & 0.40±0.49 // 0.59±0.49 & 0.55±0.50 // 0.65±0.48 & 0.63±0.48 // 0.74±0.44 \\
    \model{llama-2-7b} & 0.36±0.48 // 0.44±0.50 & 0.39±0.49 // 0.47±0.50 & 0.39±0.49 // 0.48±0.50 \\
    \model{gemma-7b} & 0.48±0.50 // 0.63±0.48 & 0.53±0.50 // 0.67±0.47 & 0.64±0.48 // 0.70±0.46 \\
    \model{aya-8b} & 0.47±0.50 // 0.65±0.48 & 0.58±0.49 // 0.64±0.48 & 0.66±0.47 // 0.72±0.45 \\
    \model{sarvam-m-24b} & 0.35±0.48 // 0.60±0.49 & 0.48±0.50 // 0.68±0.47 & 0.64±0.48 // 0.77±0.42 \\
    \model{krutrim-2-12b} & 0.50±0.50 // 0.62±0.49 & 0.56±0.50 // 0.71±0.46 & 0.68±0.47 // 0.73±0.45 \\
    \model{nanda-10b} & 0.36±0.48 // 0.48±0.50 & 0.42±0.49 // 0.53±0.50 & 0.43±0.50 // 0.56±0.50 \\
    \model{aryabhatta-8b} & 0.38±0.48 // 0.47±0.50 & 0.40±0.49 // 0.49±0.50 & 0.42±0.49 // 0.57±0.50 \\
    \model{airavata-7b} & 0.36±0.48 // 0.38±0.49 & 0.37±0.48 // 0.38±0.49 & 0.41±0.49 // 0.42±0.49 \\
    \model{gajendra-7b} & 0.37±0.48 // 0.40±0.49 & 0.37±0.48 // 0.41±0.49 & 0.39±0.49 // 0.46±0.50 \\
    \bottomrule
    \end{tabular}
    \caption{Accuracy (mean and standard deviation) on NormAd under all prompting contexts (Country, Value, Rule-of-Thumb). Entries show accuracy on Indian (left) and American (right) norms.}
    \label{tab:normad_raw_results}
\end{table*}

Accuracy results for the NormAd dataset shown in Figure~\ref{fig:normad} are reported in Table~\ref{tab:normad_raw_results}.

\section{Reproducibility Details} \label{appendix:reproducibility}

\subsection{HuggingFace Model Identifiers}
\begin{table}[h!]
\centering
\begin{tabular}{|l|l|c|}
\hline
\textbf{Model Identifier} & \textbf{HuggingFace Identifier} & \textbf{Release Year} \\
\hline
\model{sarvam-m-24b}     & sarvamai/sarvam-m                                & May 2025 \\
\model{nanda-10b}        & MBZUAI/Llama-3-Nanda-10B-Chat                   & Apr 2025 \\
\model{krutrim-2-12b}    & krutrim-ai-labs/Krutrim-2-instruct              & Feb 2025 \\
\model{airavata-7b}      & ai4bharat/Airavata                               & Jan 2024 \\
\model{aryabhatta-8b}    & GenVRadmin/AryaBhatta-GemmaOrca-Merged          & Apr 2024 \\
\model{gajendra-7b}      & BhabhaAI/Gajendra-v0.1                           & Feb 2024 \\
\model{mistral-3.1-24b}  & mistralai/Mistral-Small-3.1-24B-Instruct-2503    & Mar 2025 \\
\model{llama-2-7b}       & meta-llama/Llama-2-7b-chat-hf                    & Jul 2023 \\
\model{llama-3.1-8b}     & meta-llama/Llama-3.1-8B-Instruct                 & Jul 2024 \\
\model{gemma-7b}         & google/gemma-7b-it                               & Feb 2024 \\
\model{aya-8b}           & CohereForAI/aya-expanse-8b                       & Oct 2024 \\
\model{gpt-4o}           & N/A (OpenAI proprietary)                         & May 2024 \\
\hline
\end{tabular}
\caption{Hugging Face identifiers\chadded{ and release years} for all models evaluated in this work.}
\label{tab:hf_identifiers}
\end{table}

We used the open-source HuggingFace checkpoints for most models; see Table~\ref{tab:hf_identifiers} for the exact HuggingFace checkpoints used. We used 8-bit quantized versions of \model{sarvam-m-24b} and \model{mistral-3.1-24b}. This is standard practice for models that are too large to fit on available GPU memory. We accessed GPT-4o from the OpenAI API, using model ID \texttt{gpt-4o-2024-08-06}.

\subsection{Decoding}
\chadded{Decoding refers to the process of sampling the next token from the model's predicted distribution. Picking the most probable token is called greedy decoding. However, greedy decoding locks a model into a single response that may not answer the question or follow the formatting instructions. To reduce invalid outputs from models with brittle instruction-following capabilities, we use random sampling with a temperature of 0.3 instead of greedy decoding. To account for this randomness, we follow best practices~\cite{Song2025Good} by repeating each prompt multiple times and averaging across trials (details in Section~\ref{subsec:eval_methodology}).}

\subsection{Handling Unanswered/Refused Questions}
Models sometimes refused to answer questions, especially on sensitive World Value Survey question topics like abortion. For the cultural map, we added an explicit instruction to prevent refusal. Further, as explained in Section~\ref{sec:models_and_prompting}, instead of greedy decoding, we sample the next token with a low temperature (0.3). This gave the model a chance to answer the question without refusal or instruction-following failures. Finally, we tried sampling a valid answer ten times before counting it as invalid. The opinion alignment task relies on logit-biasing without sampling a token, and hence is safe from refusal.

\subsection{Hardware Details}
\chadded{Experiments were run on Google Cloud Platform using a combination of Nvidia T4 and A100 GPUs.}

\end{document}